\documentclass[conference]{IEEEtran}
\IEEEoverridecommandlockouts

\usepackage{cite}
\usepackage{amsmath,amssymb,amsfonts}
\usepackage{luatex85}
\usepackage{newtxtext}
\usepackage{algorithmic}
\usepackage{graphicx}
\usepackage{textcomp}
\usepackage{xcolor}
\usepackage[utf8]{inputenc} 
\usepackage[T1]{fontenc}    
\usepackage{hyperref}       
\usepackage{url}            
\usepackage{booktabs}       
\usepackage{amsfonts}       
\usepackage{nicefrac}       
\usepackage{microtype}      
\usepackage{gensymb}
\usepackage{multirow}
\usepackage{array}
\usepackage{rotating}
\usepackage{tabularx}
\usepackage{float}
\usepackage{placeins}
\usepackage{enumitem}
\usepackage{subfig}  
\usepackage{longtable}

\def\BibTeX{{\rm B\kern-.05em{\sc i\kern-.025em b}\kern-.08em
    T\kern-.1667em\lower.7ex\hbox{E}\kern-.125emX}}

\usepackage{fancyhdr}
\newcommand{\github}{\url{https://github.com/ROC-HCI/Hi5}}

\newcommand{\masum}[1]{\textcolor{black}{#1}}

\begin{document}

\title{\textbf{Hi5} \includegraphics[height=1em]{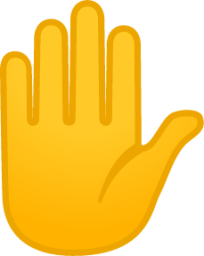}: Synthetic Data for Inclusive, Robust, Hand Pose Estimation}

\author{
\IEEEauthorblockN{
Masum Hasan, 
Cengiz Ozel, 
Nina Long\IEEEauthorrefmark{1}, 
Alexander Martin\IEEEauthorrefmark{1}, 
Samuel Potter\IEEEauthorrefmark{1}, \\
Tariq Adnan, 
Sangwu Lee, 
Ehsan Hoque\IEEEauthorrefmark{2}
}
\IEEEauthorblockA{
Department of Computer Science, University of Rochester\\
Rochester, NY, USA\\
Ministry of Defense, KSA\IEEEauthorrefmark{2}\\
Emails: m.hasan@rochester.edu, mehoque@cs.rochester.edu
}
\thanks{\IEEEauthorrefmark{1}Equal contributions, ordered by surname}
}

\maketitle
\thispagestyle{fancy}

\begin{abstract}
Hand pose estimation plays a vital role in capturing subtle nonverbal cues essential for understanding human affect. However, collecting diverse, expressive real-world data remains challenging due to labor-intensive manual annotation that often underrepresents demographic diversity and natural expressions. To address this issue, we introduce a cost-effective approach to generating synthetic data using high-fidelity 3D hand models and a wide range of affective hand poses. Our method includes varied skin tones, genders, dynamic environments, realistic lighting conditions, and diverse naturally occurring gesture animations. The resulting dataset, Hi5, contains 583,000 pose-annotated images, carefully balanced to reflect natural diversity and emotional expressiveness. Models trained exclusively on Hi5 achieve performance comparable to human-annotated datasets, exhibiting superior robustness to occlusions and consistent accuracy across diverse skin tones -- which is crucial for reliably recognizing expressive gestures in affective computing applications. Our results demonstrate that synthetic data effectively addresses critical limitations of existing datasets, enabling more inclusive, expressive, and reliable gesture recognition systems while achieving competitive performance in pose estimation benchmarks. The Hi5 dataset, data synthesis pipeline, source code, and game engine project are publicly released to support further research in synthetic hand-gesture applications.\footnote{\github}

\end{abstract}


\section{Introduction}

\begin{figure*}[ht]
\centering
\subfloat[]{
    \includegraphics[width=0.23\textwidth]{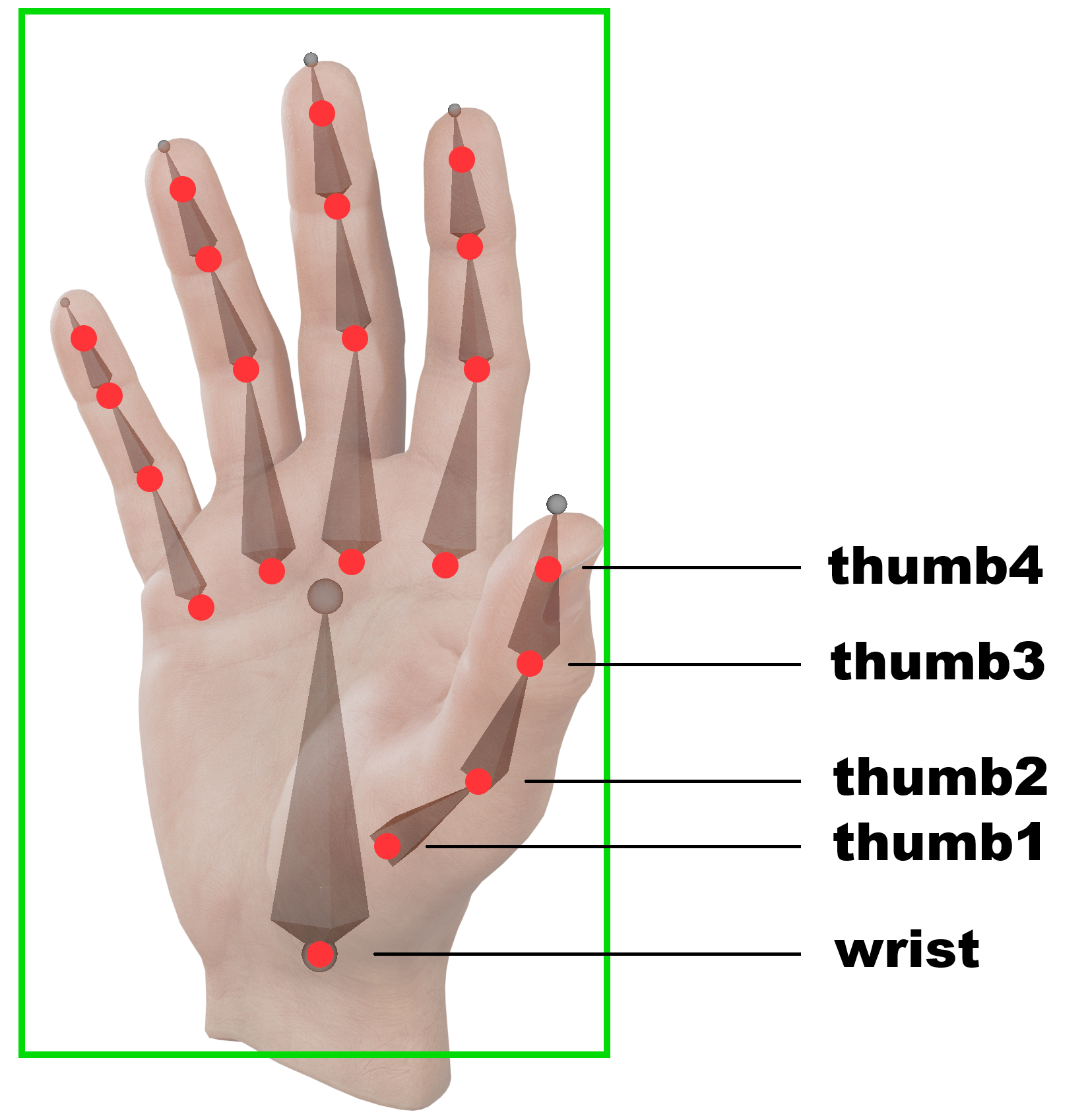}
    \label{fig:sub-a}
}
\hfill
\subfloat[]{
    \includegraphics[width=0.73\textwidth]{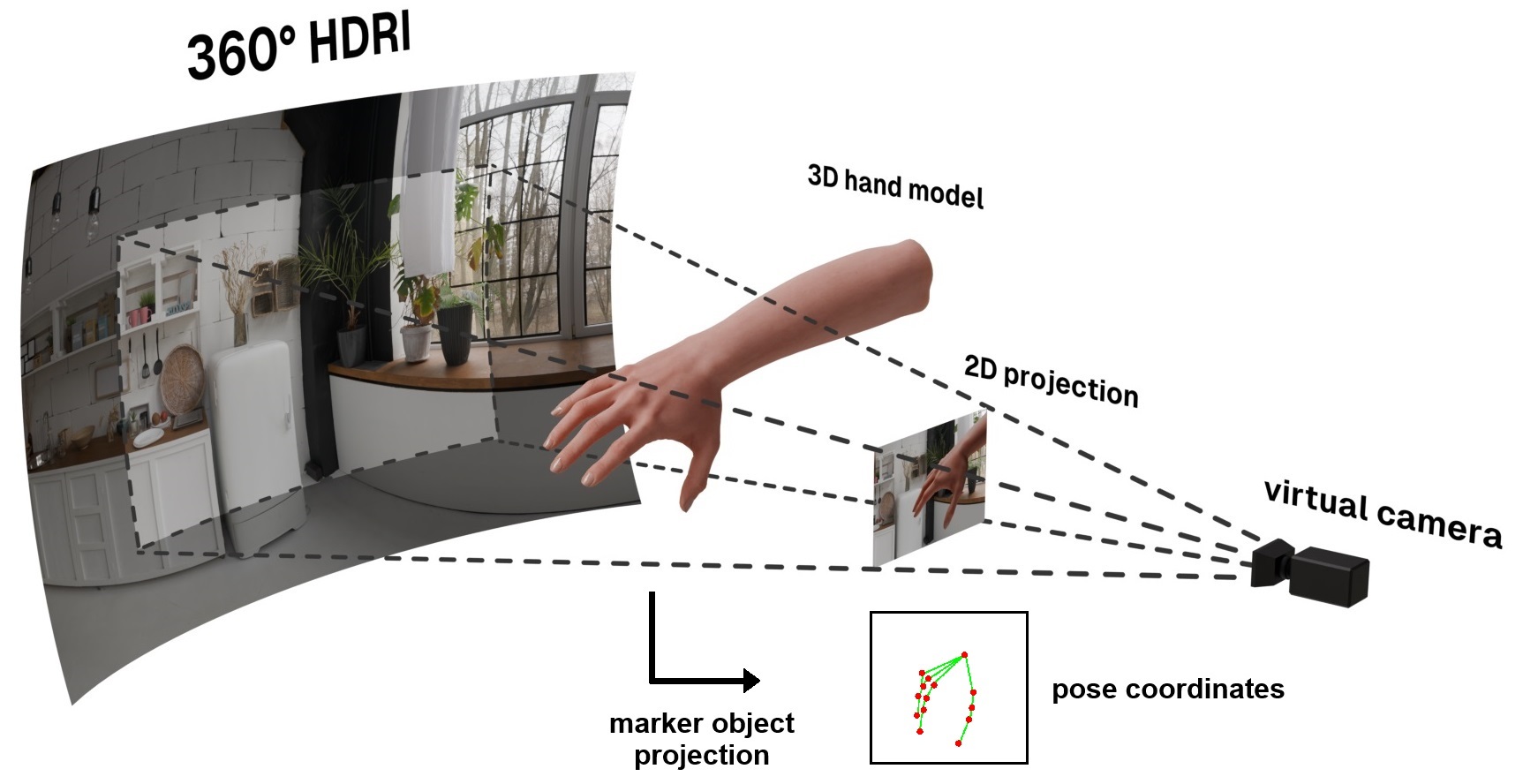}
    \label{fig:sub-b}
}
\caption{a) Invisible marker objects (visualized with red dots) within a fully rigged 3D hand armature \masum{capable of making natural hand expressions}, and b) the projection of a realistic 3D hand model into a 2D scene. This projection enables automatic, precise, and consistent pose labeling—a key step in generating synthetic data that captures natural hand configurations, even under occlusion.}
\label{fig:full-pipeline}
\end{figure*}

\masum{Hands are not only essential tools for interacting with technology but also powerful conveyors of emotion and social intent (e.g., a clenched fist conveys anger, a trembling hand conveys fear, open palms can signal warmth, etc.). In both everyday communication and human-computer interaction, hand gestures carry emotive cues that help interpret a person’s emotional state \cite{levy,KRAUSS1996389,kelly}.} Early explorations in gesture-based interfaces \cite{videoplace, put-that-there} laid the groundwork for systems that now support applications ranging from augmented and virtual reality \cite{10.1145/3678532, 10.1145/3569499, vr1, vr2, vr3, vr4} and medical diagnostics \cite{park1, park2, pmlr-v68-jaroensri17a} to sign language translation \cite{sign1, sign2, sign3, 10.1145/3610881}. \masum{More recently, the emerging field of affective computing has underscored the role of expressive hand gestures in conveying subtle emotional states and fostering embodied interactions \cite{hand-over-face,BLYTHE2023105260,luo,chua2024motion}.}

However, despite significant progress in computer vision and gesture recognition, state-of-the-art hand pose estimation models often struggle under challenging real-world conditions \masum{necessary for robust affect recognition} -- such as occlusions and variations in skin tone. This limitation is partly due to the shortcomings of popular datasets, which are typically gathered in controlled laboratory environments \cite{nyuhands, interhands, zimmermann2019freihand} or in-the-wild scenarios with inconsistent representation \cite{onehand10k}. \masum{Such limitations can impair a system’s ability to correctly interpret nonverbal cues, thereby affecting fairness and reliability. Moreover, manual annotation of hand poses is labor-intensive, error-prone, and often fails to capture the full diversity of natural, emotion-infused gestures.}

To address these challenges, we propose a novel approach for generating a diverse, representative, and large-scale synthetic hand pose dataset using only a consumer-grade computer. Our method leverages high-fidelity 3D hand models with \masum{a large set of day-to-day realistic hand animations}, varied genders, and multiple skin colors, combined with dynamic environmental and lighting conditions. This synthetic data generation process naturally spans a wide range of views and angles—including those typical in affective, embodied interactions where hands may be partially occluded or in unconventional positions. Unlike human annotation, our pipeline produces consistent labels even for invisible parts of the hand, thereby enhancing robustness against occlusion.

We introduce Hand Images 500K (Hi5), a synthetic dataset comprising 583,000 realistically rendered hand images in natural environments. The dataset is carefully balanced with equal representation of two genders and six representative skin colors based on the Dermatological ITA scale~\cite{ita}. \masum{We captured 44 different commonly used hand expressions and interpolated them to create novel hand animations to introduce variability to the dataset.} Models trained solely on Hi5 not only perform competitively with those trained on real data but also exhibit enhanced robustness in challenging conditions such as occlusions and varied skin tones.

Our contributions are threefold:
\begin{enumerate} 
    \item We introduce a novel synthetic data synthesis pipeline that provides precise control over the diversity and representation of hand poses, capturing naturalistic variations and \masum{expressive movements} without human annotation.
    \item We present Hi5, a large-scale synthetic hand pose estimation dataset generated with consumer-grade hardware with \masum{a wide range of common human hand gestures, that surpasses existing datasets in both size and diversity.}
    \item We demonstrate that models trained solely on Hi5 achieve competitive performance on real-world benchmarks, validating synthetic data as a practical alternative for developing robust, emotion-aware gesture recognition systems.
\end{enumerate}

In affective computing and embodied interaction, high-quality hand tracking is a key enabler for interpreting subtle emotional signals and creating naturalistic digital interfaces. By mitigating the biases and limitations of manually annotated datasets, our approach opens new avenues for applications that require fair, robust, and reliable gesture recognition for all people in any environment. In the following sections, we review related work (Section 2), describe the Hi5 dataset generation process (Section 3), outline our training and evaluation methods (Section 4), present experimental results (Section 5), discuss insights and future directions (Section 6), and conclude (Section 7). Alongside this paper, we release our data synthesis pipeline, source code, and the 583K pose-annotated synthetic image dataset to support future research in gesture-based interaction systems.
\section{Related Work}

\masum{Hands play an important role in conveying emotions and intentions in human interactions, and accurate hand pose estimations provide an accessible, low-cost way to recognize gestures and granular affects. This section reviews key literature on hand gesture for affective computing, and existing techniques and datasets for hand pose estimation.}

\subsection{Hand Gesture in Affective Computing}

\masum{Hand gestures have been strongly related to affect. Mahmoud et al. \cite{hand-over-face} found that face occlusion with hands serves as a cue for recognizing cognitive and mental states. Levy et al. \cite{levy} showed that hand gestures of a speaker significantly enhance memory retention for spoken sentences. Complementing this finding, Krauss et al. \cite{KRAUSS1996389} argued that conversational hand gestures primarily assist in the cognitive planning and production of speech, thus supporting communicative processes during interactions. Additionally, Kelly et al. \cite{kelly} highlighted the dual function of hand gestures--not only facilitating language acquisition, speech formulation, conceptual learning, problem-solving, meaning comprehension, and the externalization of thought, but also enabling speakers to convey emotions effectively. These gestures particularly aid listeners in emotionally interpreting speech through metaphoric, expressive, and multimodal cues.}

\masum{Blythe et al. \cite{BLYTHE2023105260} demonstrated that emotions can be accurately perceived from isolated body parts, with hands identified as particularly informative compared to arms, heads, or torsos. Even a single hand provided substantial emotional insight, highlighting the critical role of hands in emotion perception. Luo et al. \cite{luo} similarly revealed that simple one-hand gestures effectively convey emotional information, noting that finger-pointing direction and gesture intensity are closely associated with emotional valence and arousal. Similarly, Chua et al. \cite{chua2024motion} found that hand gesture features like speed, distance, and tension can reveal a user's emotional state and cognitive load in VR. Using only gesture data, support vector classifiers accurately predicted high or low levels of valence, arousal, and cognitive load, suggesting gestures alone can be used for emotion and workload detection without extra sensors.}

\subsection{Existing Methods and Datasets for Hand Pose Estimation}

Hand pose estimation methods, \masum{while foundational for accessible vision-based interpretation of hand gestures}, face critical limitations in recognizing affect through hand modality in real-world scenarios. Though deep learning frameworks like CNNs (e.g., DeepPose \cite{deeppose}) and transformers (e.g., ViTPose \cite{xu2022vitpose}) have advanced body pose estimation, their adaptation to hand pose estimation remains constrained by limited datasets and added challenges with tracking 21 hand keypoints (instead of 17 in human body) across highly articulated gestures. These challenges that are common in unconstrained environments, which are compounded by occlusions, low lighting, and dynamic interactions. Early depth sensor-dependent approaches (e.g., Qian et al. \cite{qian2014hand}, Tompson et al. \cite{tompson2014real}, Sinha et al. \cite{sinha2016deephand}) limit accessibility for in-the-wild RGB-based affect analysis.

Existing datasets further undermine affect recognition capabilities. Real-world datasets like NYU Hand Pose \cite{tompson14tog} and CMU Panoptic HandDB \cite{simon2017hand}, captured in controlled settings, lack the environmental, demographic, and cultural diversity needed to model nuanced affect across populations. FreiHAND \cite{zimmermann2019freihand} improves gender/racial diversity but introduces synthetic backgrounds with unrealistic lighting, failing to capture naturalistic contexts where emotional gestures occur. Synthetic datasets (e.g., SynthHands \cite{OccludedHands_ICCV2017}, GANerated \cite{mueller2018ganerated}) lack photorealism and balanced skin-tone representation, biasing models toward homogeneous populations. In-the-wild datasets like OneHand10K \cite{onehand10k} include environmental variability but overlook demographic balance. Crucially, none of the existing work prioritizes scenarios essential for affect recognition: diverse and natural hand poses, occlusions, low-light settings, or balanced gender and skin tone representation.

\masum{Hand gesture is a critical modality for conveying affect, yet existing pose estimation methods and datasets remain inadequate for robust real-world emotion recognition. Our paper aims to fill this crucial gap.}

\section{Pose-Annotated Synthetic Hand Image Generation}

Figure \ref{fig:full-pipeline} demonstrates an overview of our data synthesis pipeline. We use a set of virtual 3D hands with diverse skin textures, and place them in realistic 3D environments with articulated and expressive poses. We take dynamic virtual pictures and use them as training data.


To simulate realistic human hands inside a 3D game engine, we purchased 2 pairs of high-fidelity 3D human hand (1 male, 1 female) models from a 3D object marketplace. The hand models are fully rigged for animation. Each model came with two skin textures: pale and darker. For rendering in a ``Physically Based Rendering'' environment, which means a lighting environment that follows real-world optics, the models include Albedo and Roughness textures, and Normal maps for surface detail. We included subsurface scattering in the 3D hand models, which is an optical phenomenon that allows the light to penetrate the surface of the skin -- allowing a seamless blending with the environment. Each hand model included a detachable arm, which allowed us to simulate different arm lengths. In our game engine setup, we only simulated the right hand. The left hand was created by mirroring half of the images during data augmentation.

We created a diverse set of animation keyframes for each of the 44 different affective hand poses (e.g. Good luck, Okay sign, Fake gun, etc.). The list of hand poses and movements, alongside their affective implications, are listed in Table \ref{tab:animations}. We interpolated between the keyframes, effectively animating the hand from one pose to another. This method allows us to capture a much wider variety of poses beyond the initial 44.

\begin{table}[!ht]
\captionsetup{justification=raggedright, singlelinecheck=false}
\caption{\masum{List of Poses and Motions used in the creation of Hi5 and their affective implications}}
\label{tab:animations}
\scriptsize
\begin{tabularx}{0.5\textwidth}{|p{2.5cm}|X|}
\hline
\textbf{Hand Pose / Motion} & \textbf{Implied Affect} \\
\hline
\multicolumn{2}{|c|}{\textbf{Static Poses}} \\
\hline
Neutral relaxed & Baseline state; calm or disengaged \\
Neutral rigid & Slight tension; discomfort or guardedness \\
Good luck & Hopeful anticipation, anxious optimism \\
Fake gun & Playful or mildly confrontational, context-dependent \\
Star Trek & Cultural symbol; peace, formality, or sci-fi fandom \\
Extended thumb & Request, openness, or informal friendliness \\
Thumb up relaxed & Casual approval, positive but informal \\
Thumb up normal & Agreement or encouragement \\
Thumb up rigid & Strong approval or assertive positivity \\
Thumb tuck normal & Passive stance; possibly insecure \\
Thumb tuck rigid & Guarded or defensive self-comfort \\
A-okay & Approval, everything is fine \\
A-okay upright & Assertive or emphatic approval \\
Surfer (shaka) & Relaxed, friendly, laid-back \\
Rocker & Enthusiasm, rebelliousness (rock culture) \\
Rocker front & Expressive engagement or intensity \\
Rocker back & Dismissive or distanced gesture \\
Fist & Determination, strength, or readiness \\
Fist rigid & Aggression, anger, or tension \\
Alligator closed & Playful or aggressive storytelling \\
One count & pointer finger \\
Two count & Peace sign, V \\
Three count & Number three \\
Four count & Number four \\
Five count & Open palm, displaying openness \\
Index tip & Precision or pointing focus \\
Middle tip & Emphasis or clarity in focus \\
Ring tip & Delicacy or gentleness \\
Pinky tip & Subtlety or playfulness \\
Palm up & Openness, offering, or request \\
Finger spread relaxed & Casual openness or readiness \\
Finger spread normal & Attention or mild alertness \\
Finger spread rigid & High tension or strong assertiveness \\
Capisce & Inquiry, emphasis, or theatrical disbelief \\
Claws & Aggression, frustration, or drama \\
Peacock & Display, pride, or flamboyance \\
Cup & Nurturing, receiving, or inquiry \\
Shakespeare's Yorick & Reflective, dramatic, sorrowful \\
Dinosaur & Playful, exaggerated, humorous \\
Middle finger & Aggression, rejection, or disrespect \\
\hline
\multicolumn{2}{|c|}{\textbf{Motions}} \\
\hline
Relaxed wave & Friendly or casual greeting/farewell \\
Fist wave & Unity or playful strength \\
Prom wave & Formality, politeness, or sarcasm \\
\hline
\end{tabularx}
\end{table}

\subsection{Automatic Image and Pose Capture}
\label{subsub:data-capture}

Consistent with prevalent models in hand pose estimation \cite{openpose-hand, mediapipehands, onehand10k}, we selected 21 anatomically relevant points on a typically-abled human hand. We inserted invisible marker objects at each one of these keypoints inside the 3D hand model as demonstrated in Figure \ref{fig:full-pipeline} (a). We rendered each frame during playback in real-time and saved the images in $640\times640$ resolution to disk at 60 frames per second. Alongside the images, we also saved the invisible marker positions (Figure \ref{fig:full-pipeline}) and frame metadata.


\subsection{Data Diversity}
\label{subsub:data-diversity}





\begin{figure}[!t]
    \centering
    \includegraphics[width=\linewidth]{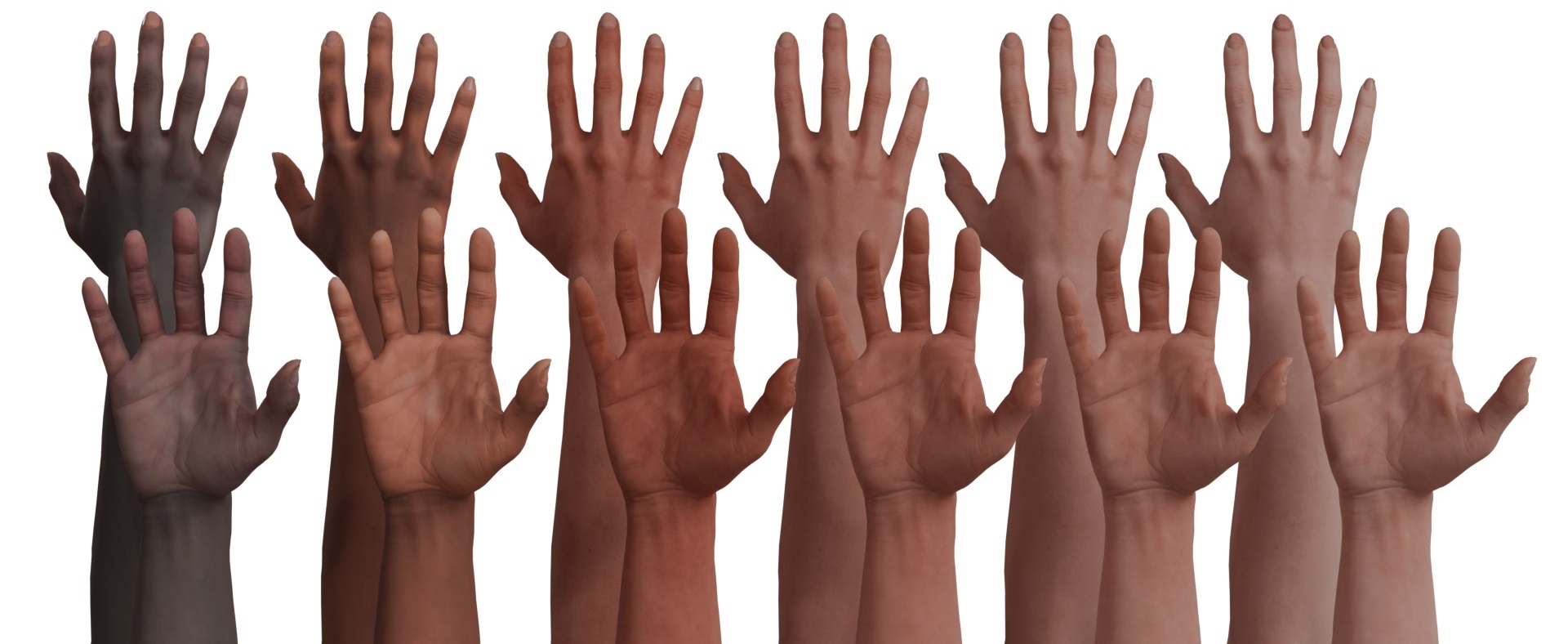}
    \caption{Our synthetic data pipeline allows us precise control of skin color representation. This figure demonstrates the female hand models used in Hi5 with skin tone ITA values respectively -80, -30, 10, 28, 41, and 55 based on dermatology literature\cite{chardon1991skin, ita}. Full ITA scale Appendix Figure \ref{fig:ita-range}.}
    \label{fig:skin-tones}
\end{figure}

\textbf{Gender and Skin Color Diversity:} 
We incorporated two distinct base hand models---male and female---to ensure gender inclusivity, alongside six meticulously selected skin tones reflective of global populations. The skin tones in our study are chosen based on Individual Typology Angle (ITA) values, a recognized dermatological scale for categorizing skin colors where higher numbers indicate brighter skin tones \cite{chardon1991skin, ita}. 

\textbf{Dynamic Environment and Lighting:}
We use 111 HDRI environments from open-access marketplaces—360\textdegree{} scans of real indoor and outdoor scenes with rich lighting data. This enables realistic lighting of our 3D hand model, enhanced further with subsurface scattering to mimic skin translucency. Each image uses a random Z-axis HDRI rotation and varied exposure to ensure diverse, lifelike conditions while maintaining high configurability.


\textbf{Camera Position and Angle:}
We randomly placed the camera around the 3D hand using polar coordinates, varying distances, and adding small random $X$, $Y$, and $Z$ axis rotations. This captures diverse views—first-, second-, and third-person—and simulates close-up to distant shots. Up to $25\%$ of the hand is allowed to be outside the frame to teach the model to predict unseen parts. All poses are marked in a consistent Cartesian coordinate system, allowing out-of-bound values.

\textbf{Data Augmentation:}  
Following \cite{wood2021fake}, we apply data augmentation to reduce domain gaps and increase diversity. Since image generation is cheap, we use in-place augmentation, replacing original images. Each image undergoes several augmentation steps with a fixed probability (Appendix Table~\ref{tab:augmentation}); $79.18\%$ of images are altered.

Our augmentations include:  
\begin{enumerate}
\item \textbf{Geometric Transformations:} Resampling, scaling, stretching, and translation with pose coordinate adjustment \cite{shorten2019survey}.
\item \textbf{Color Space Operations:} Brightness, balance, contrast, and histogram edits to improve lighting robustness \cite{zoph2020learning}.
\end{enumerate}
Other methods include blurring, flipping (left/right hand balance), and Gaussian erase, which hides random hand regions to improve generalization.





\begin{figure}[!ht]
    \centering
    \begin{minipage}{\linewidth}
    \centering
    \setlength{\tabcolsep}{2pt}
    \renewcommand{\arraystretch}{0.5}
    \begin{tabular}{ccc}
        \makebox[0.30\textwidth][c]{\begin{tabular}{c}
            \includegraphics[width=0.30\textwidth]{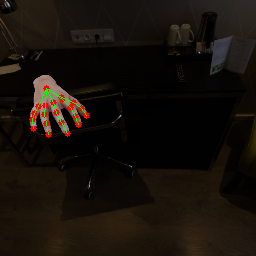} \\
            \small Finger spread relaxed
        \end{tabular}} & 
        \makebox[0.30\textwidth][c]{\begin{tabular}{c}
            \includegraphics[width=0.30\textwidth]{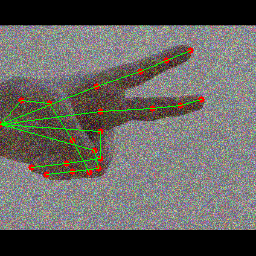} \\
            \small Two count (peace)
        \end{tabular}} & 
        \makebox[0.30\textwidth][c]{\begin{tabular}{c}
            \includegraphics[width=0.30\textwidth]{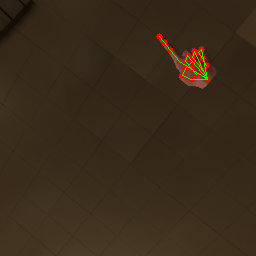} \\
            \small One count (point)
        \end{tabular}} \\
        \makebox[0.30\textwidth][c]{\begin{tabular}{c}
            \includegraphics[width=0.30\textwidth]{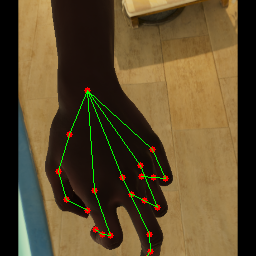} \\
            \small Claws
        \end{tabular}} & 
        \makebox[0.30\textwidth][c]{\begin{tabular}{c}
            \includegraphics[width=0.30\textwidth]{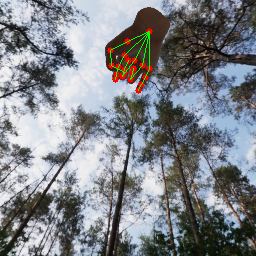} \\
            \small Fist
        \end{tabular}} & 
        \makebox[0.30\textwidth][c]{\begin{tabular}{c}
            \includegraphics[width=0.30\textwidth]{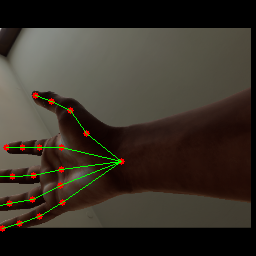} \\
            \small Palm up
        \end{tabular}} \\
        \makebox[0.30\textwidth][c]{\begin{tabular}{c}
            \includegraphics[width=0.30\textwidth]{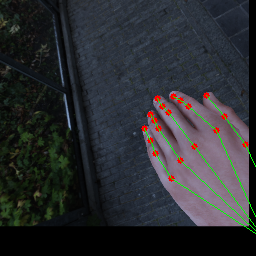} \\
            \small Neutral rigid
        \end{tabular}} & 
        \makebox[0.30\textwidth][c]{\begin{tabular}{c}
            \includegraphics[width=0.30\textwidth]{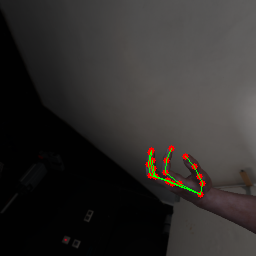} \\
            \small Pointer tip
        \end{tabular}} & 
        \makebox[0.30\textwidth][c]{\begin{tabular}{c}
            \includegraphics[width=0.30\textwidth]{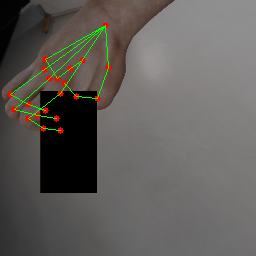} \\
            \small Pinky tip
        \end{tabular}} \\
    \end{tabular}
    \end{minipage}
    \caption{Sample of different affective hand pose images from the Hi5 dataset with automatically generated pose labels. Affect labels are estimations, as different poses interpolate between each other.}
    \label{fig:synthesis}
\end{figure}

\begin{figure}[!ht]
    \centering
    \begin{minipage}{\linewidth}
    \centering
    \setlength{\tabcolsep}{2pt}
    \renewcommand{\arraystretch}{0.5}
    \begin{tabular}{ccc}
        \includegraphics[width=0.30\textwidth]{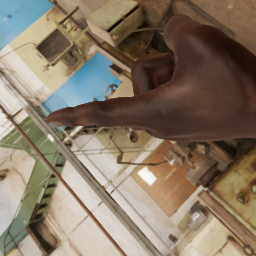} & 
        \includegraphics[width=0.30\textwidth]{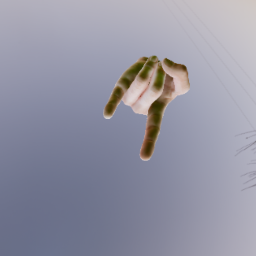} & 
        \includegraphics[width=0.30\textwidth]{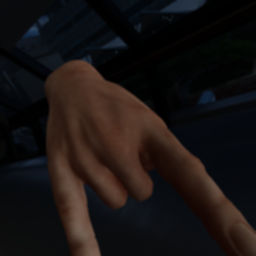} \\
        \includegraphics[width=0.30\textwidth]{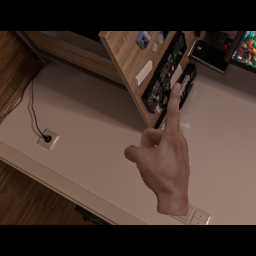} & 
        \includegraphics[width=0.30\textwidth]{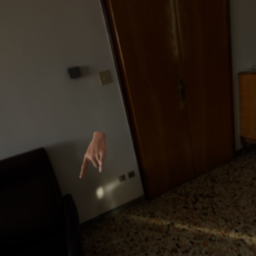} & 
        \includegraphics[width=0.30\textwidth]{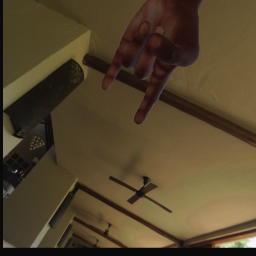} \\
    \end{tabular}
    \end{minipage}
    \caption{Nearly identical hand poses in our dataset have a surprisingly diverse image representation. (Pose: Rocker)}
    \label{fig:same-pose}
\end{figure}

We create 3 different sizes of synthetic datasets: Hi5-Large (538,643 images), Hi5-Medium (100,000 images), and Hi5-Small (10,000 images). Figure \ref{fig:synthesis} shows some sample images created through our data synthesis pipeline alongside their pose labels and their affect. Due to the randomness in our data creation process, the generated images are highly diverse. For example, Figure \ref{fig:same-pose} displays five hand images all displaying a Rocker pose. However, due to the stochasticity in our data creation process, the end images look drastically different from each other.

\section{Training \& Evaluation}
\label{sec:eval}





\subsection{Training Setup}
\label{subsec:train}
This paper aims to demonstrate the effectiveness of our synthetic data creation method on commonly used neural architectures. Therefore, we chose ViTPose \cite{xu2022vitpose}, a simple yet effective pose estimation training framework on top of a non-hierarchical vision transformer (ViT) \cite{vit}. ViTPose achieved state-of-the-art in multiple pose estimation benchmarks while being efficient to train. ViTPose appends several simple decoder layers after the pretrained vision transformer backbone to predict the pose estimation. This takes advantage of the generic vision capabilities of the pretrained vision transformer and translates that to pose estimation. We chose the ViT-Small model trained with masked autoencoder (MAE) \cite{mae} as our training backbone as it is lightweight and easy to train.

We train 4 instances of the same ViTPose Small model following the official implementation\footnote{\url{https://github.com/ViTAE-Transformer/ViTPose}}. First, with each of the three different sizes of synthetic dataset: Hi5-Large (538,643 images), Hi5-Medium (100,000 images), Hi5-Small (10,000 images), then one human-annotated hand pose estimation dataset: OneHand10K (10,000 images). In training each model, the checkpoint with the best AUC in the validation set is saved. Each model is trained for a maximum of 400 epochs and stopped early if the performance plateaus.

\subsection{Evaluation on Real Data}
\label{subsec:eval}
ViTPose follows the common top-down setting for pose estimation, which predicts the pose coordinates given the object (e.g. left and right hand) given the object location using a separate detector model. For our training and evaluation, we use the bounding box data from the ground truth, and similar to the original ViTPose paper\cite{xu2022vitpose} we evaluate the models on pose estimation performance only.

In this section, we evaluate how the models trained with synthetic data perform hand pose estimation on real data, compared to a model trained with real data. For evaluating the model, we use the following metrics,
\begin{itemize}[label={},left=-5pt, itemsep=1pt, topsep=4pt]
    \item \textbf{Percentage of Correct Keypoints (PCK)} measures the proportion of correctly predicted keypoints within a certain threshold distance proportionate to the bounding box. We use, $threshold = 0.2$.
    \item \textbf{Area Under the Curve (AUC)} calculates the PCK for various thresholds and then computes the area under the resulting curve by averaging these PCK values.
    \item \textbf{End-Point Error (EPE)} is the average Euclidean distance of predicted and ground-truth keypoint in number of pixels.
\end{itemize}

\subsubsection{Real Data Benchmark}
We take the best model checkpoint from each training discussed in Subsection \ref{subsec:train} and evaluate them on OneHand10K test dataset. OneHand10K test dataset contains 1,703 in-the-wild hand gesture images and human annotation of the pose coordinates. As OneHand10K train and test data are splits from the same data distribution and follow the same annotation scheme by the same annotators, they have a natural advantage to get a high score. However, performing reasonably well in this test set gives us a validation for the effectiveness of the synthetic data. Table \ref{tab:performance_metrics} shows the performance comparison.

\subsubsection{Perturbation Test}
To evaluate robustness and real-world occlusions, we designed a challenging perturbation test set. Our data synthesis pipeline can simulate labels for part of the image that is corrupted, out of frame, or not visible. This enables a model trained with our synthetic data to be more robust to occlusion, noise, or other disturbances. To test this, we perturb the test dataset by deleting exactly half of the hand in every image in the OneHand10k test dataset. Figure \ref{fig:perturbed} in the Appendix shows examples of the perturbed test dataset. In this test, we keep the label the same as the original images, this challenges each model to predict the full hand pose by only observing half of the hand. Table \ref{tab:performance_metrics} also shows the result of this perturbation.

\subsubsection{Evaluation on Different Skin Colors}
Representation of different skin colors is a major limitation of many computer vision datasets related to humans. However, our synthetic data creation guarantees equal representation of skin colors. We would like to test our model's capability on different skin colors, particularly on darker skin colors which are rare in real datasets. In our observation, darker skin color hands are noticeably underrepresented in OneHand10k train and test dataset. Furthermore, the dataset does not come with a skin color label. Hence, for this test, we use 11k Hands dataset\cite{11kHands} that contains hand images alongside their skin color, gender, and other biometric labels. The dataset contains 4 categories for skin color with varied representation: \textit{Dark, Medium, Fair, Very Fair} (Appendix Figure \ref{fig:11k-skin-color}). We sample some images of each skin color category and create separate test sets. The 11K Hands dataset, however, does not come with hand pose labels. To alleviate this problem, we use MediaPipe\cite{mediapipe, mediapipehands}, a popular hand pose estimation library developed by Google, to extract pose estimation prediction for the images and use this data as ground truth. To make a comparison of the effectiveness of MediaPipe, we also test on OneHand10K test images with MediaPipe predictions as ground truth. The results are shown in Table \ref{tab:skin-color-test}. Note that, MediaPipe did not recognize 300 of the images in OneHand10k test set, reducing the test set size to 1403.

\section{Results}
\label{sec:results}

\begin{figure*}[ht]
    \centering
    \hspace{-1.2cm} 
    \begin{tabular}{>{\raggedleft\arraybackslash}m{1cm}*{6}{@{\hspace{2mm}}m{0.152\textwidth}@{}}}
        \begin{sideways}\textbf{Hi5-Large}\end{sideways} &
        \includegraphics[width=\linewidth]{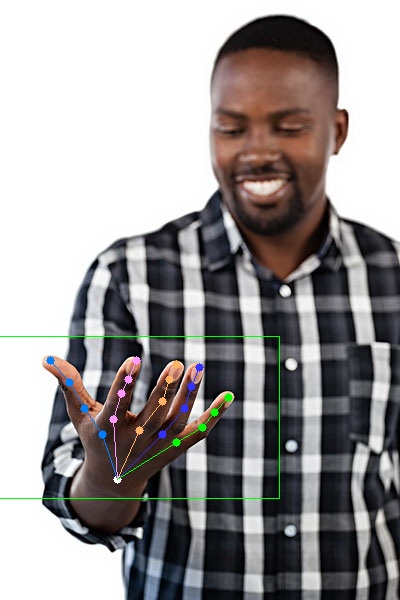} & 
        \includegraphics[width=\linewidth]{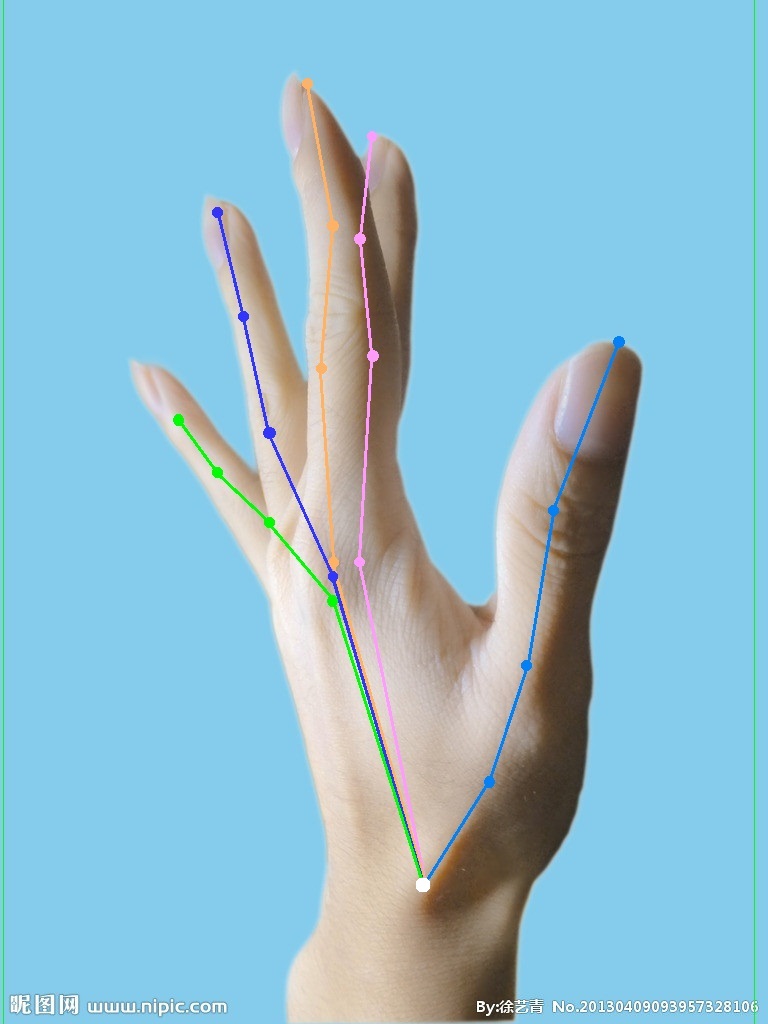} & 
        \includegraphics[width=\linewidth]{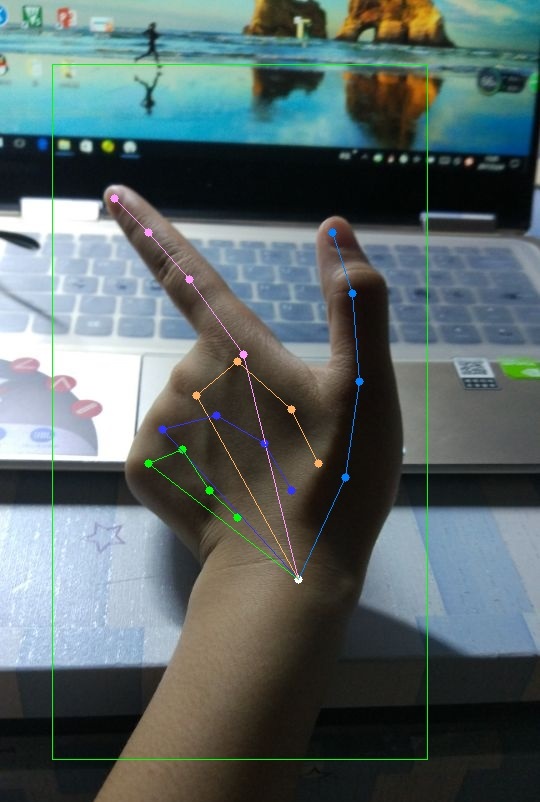} & 
        \includegraphics[width=\linewidth]{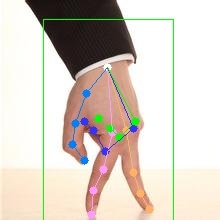}  & 
        \includegraphics[width=\linewidth]{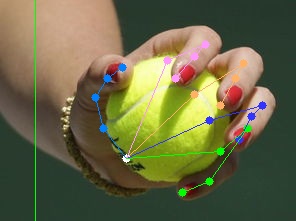} & 
        \includegraphics[width=\linewidth]{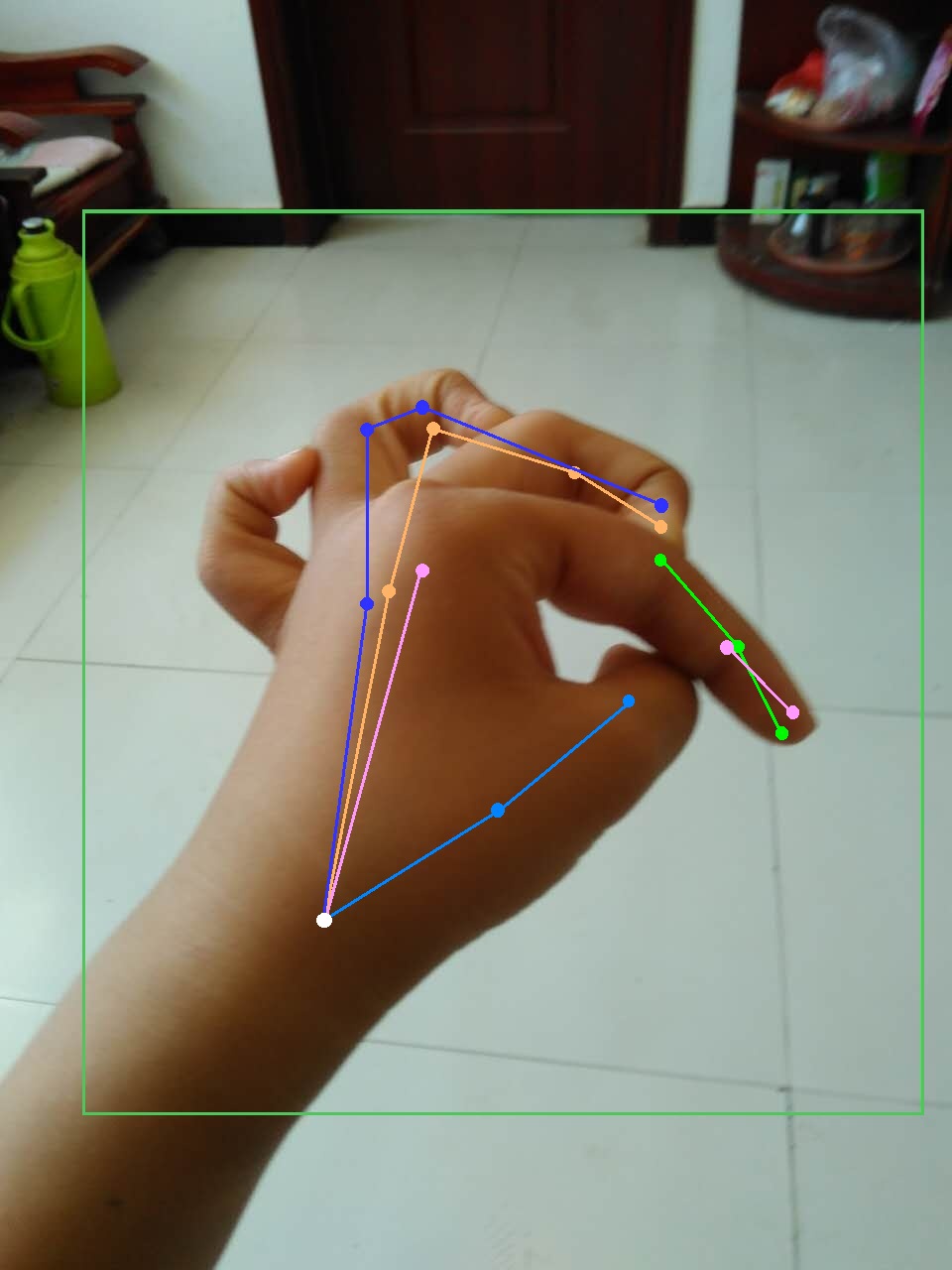}  \\[5pt]
        
        \begin{sideways}\textbf{OneHand10K}\end{sideways} &
        \includegraphics[width=\linewidth]{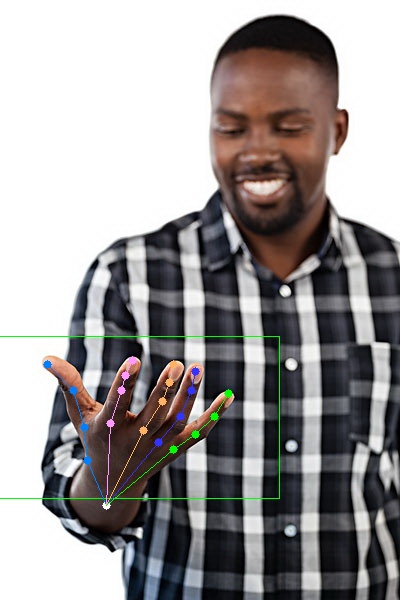} & 
        \includegraphics[width=\linewidth]{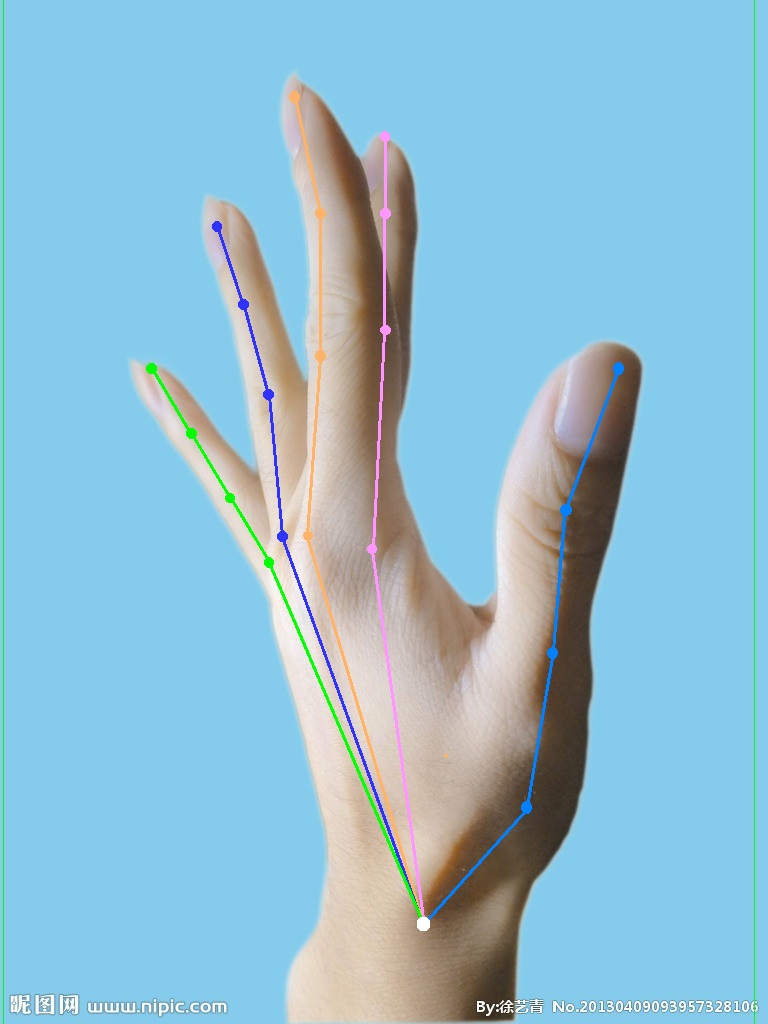} & 
        \includegraphics[width=\linewidth]{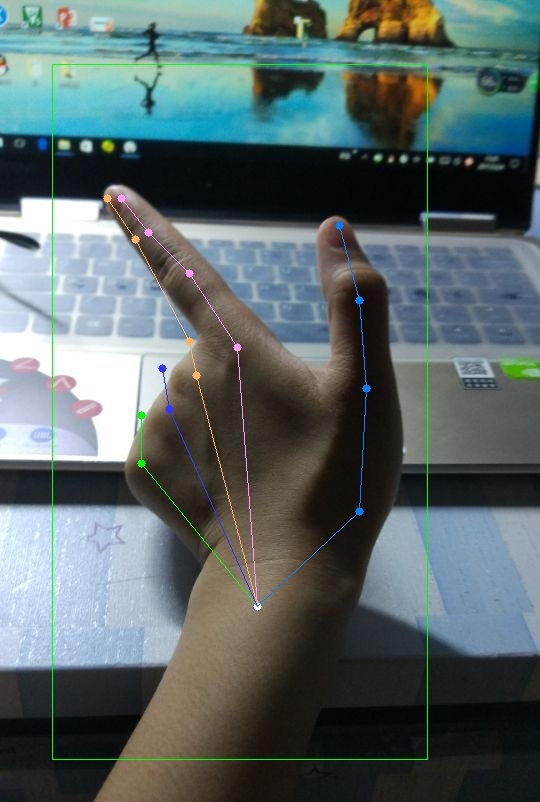} & 
        \includegraphics[width=\linewidth]{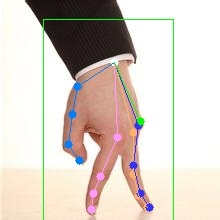} & 
        \includegraphics[width=\linewidth]{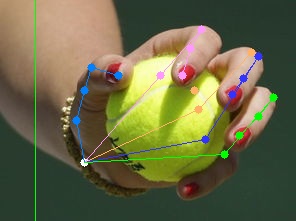} & 
        \includegraphics[width=\linewidth]{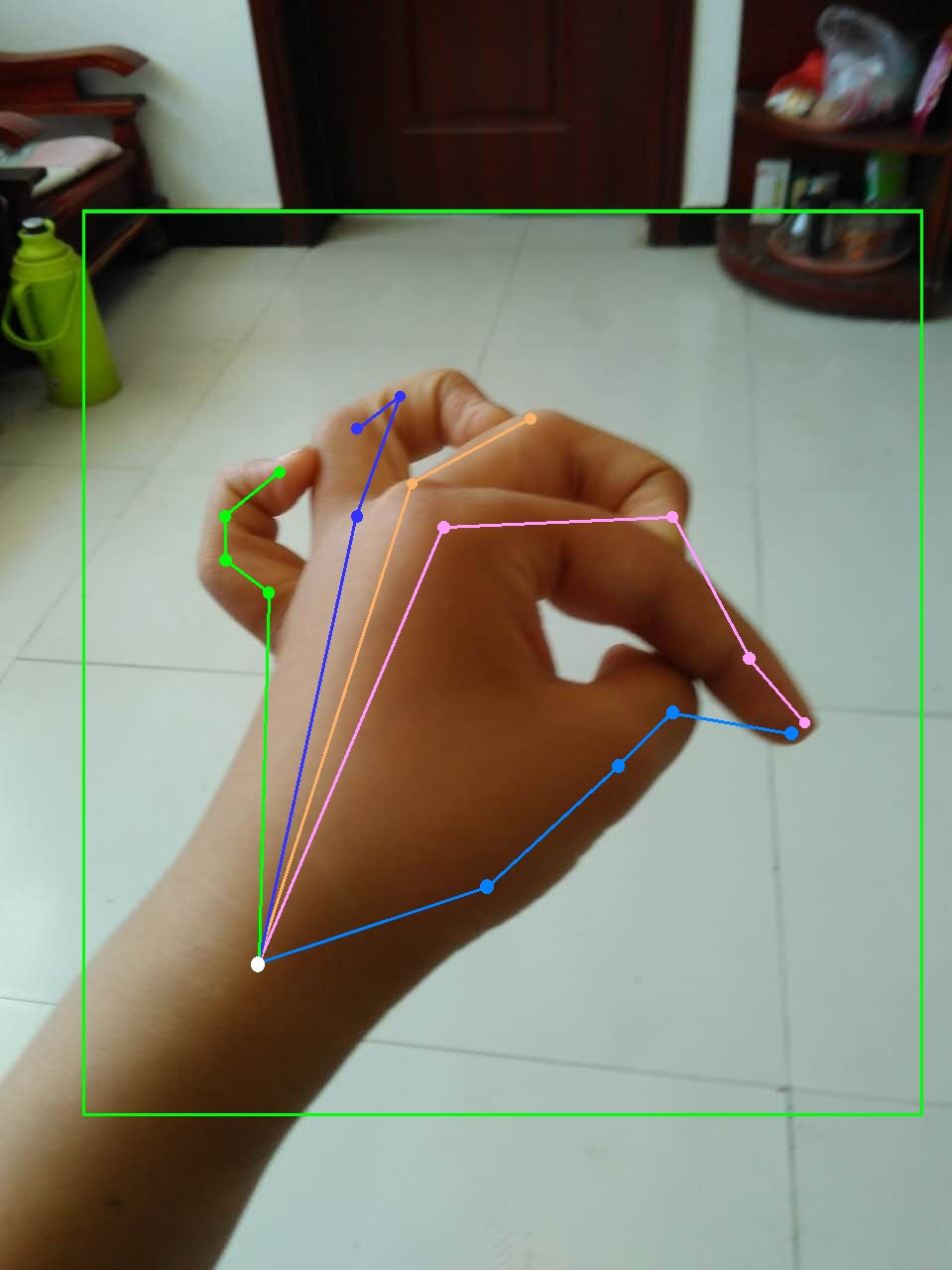} \\[5pt]
        & \centering (a) & \centering (b) & \centering (c) & \centering (d) & \centering (e) & \centering (f) \\[5pt]
    \end{tabular}
    \caption{Visual results of predictions by ViT Pose Small model trained with Hi5 Large and OneHand10K dataset  (best viewed on color and zoomed in).}
    \label{fig:visual-results}
\end{figure*}

\subsection{Quantitative Results}
Table \ref{tab:performance_metrics} demonstrates the \textit{AUC, EPE,} and \textit{PCK} performance of ViTPose-S\cite{xu2022vitpose} model trained with a real human-annotated dataset (OneHand10K) and multiple sizes of synthetic datasets (Hi5-Small, Hi5-Medium, Hi5-Large) tested on OneHand10K test set, and Perturbed OneHand10K test set. We can see in the regular test set, the model trained with OneHand10K dataset performs better in all metrics, which is expected as both the training and test data came from the same distributions of human annotation scheme. However, model trained with Hi5-Large is able to achieve a closer score while having a completely different annotation scheme and being trained entirely with synthetic data. On the other hand, the performance improvement on synthetic data with the increase in training dataset size is notable. This hints that an even larger dataset may be able to close the gap with the model trained with real data. The discrepancy in the sizes of real and synthetic dataset implies that there might be a data distillation method that could preserve the model performance with a smaller-size synthetic dataset.

Although all models suffer due to the perturbation of the data, the model trained with Hi5-Large suffers less and achieves the best results in all categories, with Hi5-Medium being a close second. This implies with significant corruption or occlusion of images, our synthetic data creates more robust models.

\begin{table*}[ht]
    \centering
    \caption{Performance Metrics on Test set and Perturbed test set}
    \begin{tabular}{lccc@{\hskip 15pt}ccc}
        \toprule
        \multirow{2}{*}{\textbf{Training Dataset}} & \multicolumn{3}{c}{\textbf{Test set}} & \multicolumn{3}{c}{\textbf{Perturbed test set}} \\
        \cmidrule(lr{15pt}){2-4} \cmidrule(lr){5-7}
         & \textbf{AUC} $\uparrow$ & \textbf{EPE} $\downarrow$ & \textbf{PCK} $\uparrow$ & \textbf{AUC} $\uparrow$ & \textbf{EPE} $\downarrow$ & \textbf{PCK} $\uparrow$ \\
        \midrule
        \noalign{\vskip 1mm}
        OneHand10K & \textbf{0.4831} & \textbf{37.6934} & \textbf{0.9856} & 0.2002 & 232.3519 & 0.6420 \\
        \noalign{\vskip 1mm}
        Hi5-Small & 0.3100 & 106.0484 & 0.8723 & 0.1450 & 246.0686 & 0.5859 \\
        \noalign{\vskip 1mm}
        Hi5-Medium & 0.3890 & 75.8657 & 0.9379 & 0.2099 & 219.0763 & 0.6887 \\
        \noalign{\vskip 1mm}
        Hi5-Large & 0.4068 & 68.0752 & 0.9552 & \textbf{0.2139} & \textbf{214.5883} & \textbf{0.6940} \\
        \noalign{\vskip 1mm}
        \hline
        \noalign{\vskip 1mm}
    \end{tabular}
    \label{tab:performance_metrics}
\end{table*}

\subsection{Skin Tone Results}
When models trained with OneHand10K (real) and Hi5-Large (synthetic) dataset are tested on hands of different skin colors, the results are mixed (Table \ref{tab:skin-color-test}). In darker hands, the Hi5-Large dataset helped achieve lower \textit{EPE} and higher \textit{PCK}, however, OneHand10K helped achieve higher \textit{AUC}. For hands of the category \textit{Very fair}, OneHand10K performs better in all metrics, which could be explained by high frequency of fair hands in the training dataset. In Table \ref{tab:skin-color-test}, when both models compared against MediaPipe\cite{mediapipe} generated pose coordinates on OneHand10K test set, their performance becomes relatively similar, with the model trained with Hi5-Large is leading in all metrics. This hints that when the advantage of the train-test same distribution and annotation scheme is taken out, synthetic data performs competitively with real data.

\begin{table*}[ht]
    \centering
    \caption{Performance metrics for different skin tones compared against MediaPipe Hands\cite{mediapipehands}.}
    \begin{tabular}{lccc@{\hskip 15pt}ccc}
        \toprule
        \multirow{2}{*}{\textbf{Test Dataset (Size)}} & \multicolumn{3}{c}{\textbf{OneHand10K}} & \multicolumn{3}{c}{\textbf{Hi5-Large}} \\
        \cmidrule(lr{15pt}){2-4} \cmidrule(lr){5-7}
         & \textbf{AUC} $\uparrow$ & \textbf{EPE} $\downarrow$ & \textbf{PCK} $\uparrow$ & \textbf{AUC} $\uparrow$ & \textbf{EPE} $\downarrow$ & \textbf{PCK} $\uparrow$ \\
        \midrule
        OneHand10K test (1403) & 0.4517 & 54.5892 & 0.9340 & \textbf{0.4585} & \textbf{52.6380} & \textbf{0.9375} \\
        \noalign{\vskip 1mm}
        Dark (635) & \textbf{0.3259} & 29.3596 & 0.9987 & 0.2970 & \textbf{25.7957} & \textbf{1.0} \\
        \noalign{\vskip 1mm}
        Medium (915)  & \textbf{0.3282} & 28.4353 & 0.9987 & 0.2978 & \textbf{26.0758} & \textbf{0.9992} \\
        \noalign{\vskip 1mm}
        Fair (939) & \textbf{0.3304} & 27.9544 & 0.9997 & 0.3003 & \textbf{25.5552} & \textbf{0.9998} \\
        \noalign{\vskip 1mm}
        Very fair (330) & \textbf{0.3953} & \textbf{21.7946} & \textbf{1.0} & 0.3228 & 23.3413 & \textbf{1.0} \\
        \noalign{\vskip 1mm}
        \hline
        \noalign{\vskip 1mm}
    \end{tabular}
    \label{tab:skin-color-test}
\end{table*}

\subsection{Qualitative Results}

Figure \ref{fig:visual-results} demonstrates sample predictions by ViTPose models trained on the Hi5-Large and OneHand10K datasets. In numerous cases, such as Figures \ref{fig:visual-results} (a), (c), (d), and (e), both models correctly predict hand poses according to their respective annotation schemes or make similar mispredictions. For example, in Figure \ref{fig:visual-results} (b), both models confuse the pointer and middle fingers as they are in an atypical order. As seen in the knuckle points of (b), the synthetic data model, shows a tendency to place keypoints in the middle of the bones similar to the invisible marker points in Figure \ref{fig:full-pipeline} (a), while human-annotated model predictions tend to stick to the surface, similar to how an annotator would label. This will allow the model trained with synthetic data a greater consistency across multiple views of a hand.

Figure \ref{fig:visual-results} (c) and (d) also demonstrate that the model trained with Hi5 can make a close reasonable approximation of the entirely invisible joints. This is a native property of synthetic data as we can generate accurate coordinates for seen or unseen joints of the hand, however, this is very difficult to capture with human annotation. As seen in (e), the model trained with Hi5 can reasonably estimate hand-object interaction, while never explicitly being trained on it. We attribute this to our data augmentation/ noise injection methods that made the model robust enough to predict through occlusion.

However, as seen in Figure \ref{fig:visual-results} (f), if the hand pose derails too far from the hand pose animations in the Hi5 dataset, the model may perform subpar the model trained with OneHand10k. This implies the importance of having comprehensive hand animations and poses in data synthesis.
\section{Discussion \& Future Work}

\masum{Our work shows that synthetic data creation is a practical option for addressing critical gaps in hand pose estimation -- a precursor to affect-aware hand gesture recognition. By capturing diverse, emotion-focused hand poses—ranging from clenched fists to open-palmed gestures—the Hi5 dataset gives models a useful platform to identify subtle nuances in nonverbal emotional cues.}

We show that generating large-scale synthetic data for hand pose estimation is achievable on a consumer-grade GPU using open-source and open-access tools. Once the pipeline is set, producing 583K (538K train, 45K test) labeled images requires about 48 hours on an NVIDIA 3090 GPU at an estimated \$4.15 electricity cost in the United States. This pipeline ensures consistent geometry across diverse camera perspectives, annotations under occlusion, balanced demographics, and natural first- and third-person viewpoints. Models trained on this dataset can handle novel poses, partial obstructions, and hand-object interactions without explicit supervision. The dataset also enforces consistent bone structures across angles, making it suitable for 3D tasks. When deliberately perturbing test images (e.g., occlusions), models trained on the larger Hi5 dataset remain more robust, suggesting synthetic data supports generalization under stress.


However, our approach has a few key limitations. Table~\ref{tab:performance_metrics} shows that synthetic data still needs large volumes to match real-data performance. Data distillation could help identify the most useful subset. Manual hand animation remains labor-intensive, and devices like Leap Motion Sensors can be inaccurate. Future research might try commercial tracking gloves to capture more complex motions. The dataset currently lacks variations in age-related skin changes, ornaments, and contextual elements like location or activity. Earlier studies such as Mueller et al.~\cite{mueller2018ganerated} used CycleGAN to refine synthetic images into more realistic ones. Future efforts can merge our game-engine approach with modern diffusion models \cite{diffusion} and image-conditioned generation \cite{controlnet} to produce highly realistic hands, environments, and interactions via text prompts to create hand pose datasets even better representative of real-life affect.

\section{Conclusion}

\masum{Our study demonstrates that synthetic data is a feasible and efficient solution for affect-aware hand pose estimation tasks.} By creating a large-scale dataset such as Hi5 with high-fidelity 3D hand models and diverse, emotion-focused animations, we show that synthetic data can serve as a robust platform for \masum{vision-based hand pose estimation, which is the precursor to recognizing gesture-based emotional cues and interactions.} Our approach allows models to generalize to real-world conditions, including occlusions and varied skin tones, while cutting down on data collection costs and time-consuming manual annotations.

Further, our results validate that synthetic hand datasets, when thoughtfully designed, ensure inclusivity across different demographics. This helps address data scarcity and fairness concerns, making it easier for the HCI and affective computing communities to develop \masum{more inclusive emotion-sensitive interfaces}. In addition, the Hi5 dataset highlights the broader potential of synthetic data to unlock a wider range of specialized, interaction-rich applications that lack large and fully annotated real-world datasets.

\section*{Ethical Impact Statement}

\masum{This work aims to improve fairness, accessibility, and inclusivity in affect-aware hand pose estimation by reducing dependency on real-world annotated datasets that often lack demographic diversity. By generating synthetic hand images with controlled variation in skin tone, gender, and affective expression, we seek to reduce bias and enhance representation, especially for underrepresented groups. Our dataset includes balanced skin tones based on dermatological scales and equal gender representation, addressing common limitations in existing benchmarks.}

\masum{All data used in this work are synthetically generated, involving no personal or biometric data from real individuals, thereby avoiding privacy concerns. However, the deployment of affect recognition systems raises broader ethical questions, including the potential for misinterpretation of emotional states, misuse in surveillance, and unintended bias if deployed without context-specific calibration. We urge researchers and developers using our datasets and models to critically evaluate downstream applications and ensure that affective technologies are deployed transparently, with clear consent, context-awareness, and accountability.}

\masum{We release the Hi5 dataset, pipeline, and code for research use under an open license to support reproducibility and encourage inclusive development practices within the community.}



\bibliographystyle{IEEEtran}
\bibliography{bib}

@ARTICLE{onehand10k,
  author={Wang, Yangang and Peng, Cong and Liu, Yebin},
  journal={IEEE Transactions on Circuits and Systems for Video Technology}, 
  title={Mask-Pose Cascaded CNN for 2D Hand Pose Estimation From Single Color Image}, 
  year={2019},
  volume={29},
  number={11},
  pages={3258-3268},
  doi={10.1109/TCSVT.2018.2879980}}

@misc{zimmermann2019freihand,
      title={FreiHAND: A Dataset for Markerless Capture of Hand Pose and Shape from Single RGB Images}, 
      author={Christian Zimmermann and Duygu Ceylan and Jimei Yang and Bryan Russell and Max Argus and Thomas Brox},
      year={2019},
      eprint={1909.04349},
      archivePrefix={arXiv},
      primaryClass={cs.CV}
}

@article{tompson14tog,
  author = {Jonathan Tompson and Murphy Stein and Yann Lecun and Ken Perlin},
  title = {Real-Time Continuous Pose Recovery of Human Hands Using Convolutional Networks},
  journal = {ACM Transactions on Graphics},
  year = {2014},
  month = {August},
  volume = {33}
}

@misc{vit,
      title={An Image is Worth 16x16 Words: Transformers for Image Recognition at Scale}, 
      author={Alexey Dosovitskiy and Lucas Beyer and Alexander Kolesnikov and Dirk Weissenborn and Xiaohua Zhai and Thomas Unterthiner and Mostafa Dehghani and Matthias Minderer and Georg Heigold and Sylvain Gelly and Jakob Uszkoreit and Neil Houlsby},
      year={2021},
      eprint={2010.11929},
      archivePrefix={arXiv},
      primaryClass={cs.CV}
}

@INPROCEEDINGS{mae,
  author={He, Kaiming and Chen, Xinlei and Xie, Saining and Li, Yanghao and Dollár, Piotr and Girshick, Ross},
  booktitle={2022 IEEE/CVF Conference on Computer Vision and Pattern Recognition (CVPR)}, 
  title={Masked Autoencoders Are Scalable Vision Learners}, 
  year={2022},
  volume={},
  number={},
  pages={15979-15988},
  doi={10.1109/CVPR52688.2022.01553}}

@misc{xu2022vitpose,
      title={ViTPose: Simple Vision Transformer Baselines for Human Pose Estimation}, 
      author={Yufei Xu and Jing Zhang and Qiming Zhang and Dacheng Tao},
      year={2022},
      eprint={2204.12484},
      archivePrefix={arXiv},
      primaryClass={cs.CV}
}

@misc{wood2021fake,
      title={Fake It Till You Make It: Face analysis in the wild using synthetic data alone}, 
      author={Erroll Wood and Tadas Baltrušaitis and Charlie Hewitt and Sebastian Dziadzio and Matthew Johnson and Virginia Estellers and Thomas J. Cashman and Jamie Shotton},
      year={2021},
      eprint={2109.15102},
      archivePrefix={arXiv},
      primaryClass={cs.CV}
}

@inproceedings{OccludedHands_ICCV2017,
 author = {Mueller, Franziska and Mehta, Dushyant and Sotnychenko, Oleksandr and Sridhar, Srinath and Casas, Dan and Theobalt, Christian},
 title = {Real-time Hand Tracking under Occlusion from an Egocentric RGB-D Sensor},
 booktitle = {Proceedings of International Conference on Computer Vision ({ICCV})},
 url = {http://handtracker.mpi-inf.mpg.de/projects/OccludedHands/},
 numpages = {10},
 month = {October},
 year = {2017}
}

@article{11kHands,

  title = {11K Hands: gender recognition and biometric identification using a large dataset of hand images},

  author = {Afifi, Mahmoud},

  journal = {Multimedia Tools and Applications},

  doi = {10.1007/s11042-019-7424-8},

  url = {https://doi.org/10.1007/s11042-019-7424-8},

  year={2019}

}

@article{chardon1991skin,
  title={Skin colour typology and suntanning pathways},
  author={Chardon, Alain and Cretois, Isabelle and Hourseau, Colette},
  journal={International journal of cosmetic science},
  volume={13},
  number={4},
  pages={191--208},
  year={1991},
  publisher={Wiley Online Library}
}

@article{shorten2019survey,
  title={A survey on image data augmentation for deep learning},
  author={Shorten, Connor and Khoshgoftaar, Taghi M},
  journal={Journal of big data},
  volume={6},
  number={1},
  pages={1--48},
  year={2019},
  publisher={Springer}
}

@inproceedings{zoph2020learning,
  title={Learning data augmentation strategies for object detection},
  author={Zoph, Barret and Cubuk, Ekin D and Ghiasi, Golnaz and Lin, Tsung-Yi and Shlens, Jonathon and Le, Quoc V},
  booktitle={Computer Vision--ECCV 2020: 16th European Conference, Glasgow, UK, August 23--28, 2020, Proceedings, Part XXVII 16},
  pages={566--583},
  year={2020},
  organization={Springer}
}

@InProceedings{interhands,
author="Moon, Gyeongsik
and Yu, Shoou-I
and Wen, He
and Shiratori, Takaaki
and Lee, Kyoung Mu",
editor="Vedaldi, Andrea
and Bischof, Horst
and Brox, Thomas
and Frahm, Jan-Michael",
title="InterHand2.6M: A Dataset and Baseline for 3D Interacting Hand Pose Estimation from a Single RGB Image",
booktitle="Computer Vision -- ECCV 2020",
year="2020",
publisher="Springer International Publishing",
address="Cham",
pages="548--564",
isbn="978-3-030-58565-5"
}

@inproceedings{mediapipe,
title	= {MediaPipe: A Framework for Perceiving and Processing Reality},author	= {Camillo Lugaresi and Jiuqiang Tang and Hadon Nash and Chris McClanahan and Esha Uboweja and Michael Hays and Fan Zhang and Chuo-Ling Chang and Ming Yong and Juhyun Lee and Wan-Teh Chang and Wei Hua and Manfred Georg and Matthias Grundmann},year	= {2019},URL	= {https://mixedreality.cs.cornell.edu/s/NewTitle_May1_MediaPipe_CVPR_CV4ARVR_Workshop_2019.pdf},booktitle	= {Third Workshop on Computer Vision for AR/VR at IEEE Computer Vision and Pattern Recognition (CVPR) 2019}}

@article{mediapipehands,
  title={Mediapipe hands: On-device real-time hand tracking},
  author={Zhang, Fan and Bazarevsky, Valentin and Vakunov, Andrey and Tkachenka, Andrei and Sung, George and Chang, Chuo-Ling and Grundmann, Matthias},
  journal={arXiv preprint arXiv:2006.10214},
  year={2020}
}

@INPROCEEDINGS {diffusion,
author = {R. Rombach and A. Blattmann and D. Lorenz and P. Esser and B. Ommer},
booktitle = {2022 IEEE/CVF Conference on Computer Vision and Pattern Recognition (CVPR)},
title = {High-Resolution Image Synthesis with Latent Diffusion Models},
year = {2022},
volume = {},
issn = {},
pages = {10674-10685},
doi = {10.1109/CVPR52688.2022.01042},
url = {https://doi.ieeecomputersociety.org/10.1109/CVPR52688.2022.01042},
publisher = {IEEE Computer Society},
address = {Los Alamitos, CA, USA},
month = {jun}
}

@INPROCEEDINGS{controlnet,
  author={Zhang, Lvmin and Rao, Anyi and Agrawala, Maneesh},
  booktitle={2023 IEEE/CVF International Conference on Computer Vision (ICCV)}, 
  title={Adding Conditional Control to Text-to-Image Diffusion Models}, 
  year={2023},
  volume={},
  number={},
  pages={3813-3824},
  doi={10.1109/ICCV51070.2023.00355}}

@article{nyuhands,
  author = {Jonathan Tompson and Murphy Stein and Yann Lecun and Ken Perlin},
  title = {Real-Time Continuous Pose Recovery of Human Hands Using Convolutional Networks},
  journal = {ACM Transactions on Graphics},
  year = {2014},
  month = {August},
  volume = {33}
}

@INPROCEEDINGS{vr1,
  author={Voigt-Antons, Jan-Niklas and Kojic, Tanja and Ali, Danish and Möller, Sebastian},
  booktitle={2020 Twelfth International Conference on Quality of Multimedia Experience (QoMEX)}, 
  title={Influence of Hand Tracking as a Way of Interaction in Virtual Reality on User Experience}, 
  year={2020},
  volume={},
  number={},
  pages={1-4},
  doi={10.1109/QoMEX48832.2020.9123085}}

@ARTICLE{vr2,
AUTHOR={Buckingham, Gavin},   
TITLE={Hand Tracking for Immersive Virtual Reality: Opportunities and Challenges},      
JOURNAL={Frontiers in Virtual Reality},      
VOLUME={2},           
YEAR={2021},      
URL={https://www.frontiersin.org/articles/10.3389/frvir.2021.728461},       
DOI={10.3389/frvir.2021.728461},      
ISSN={2673-4192}
}

@INPROCEEDINGS{vr3,
  author={Cameron, Charles R. and DiValentin, Louis W. and Manaktala, Rohini and McElhaney, Adam C. and Nostrand, Christopher H. and Quinlan, Owen J. and Sharpe, Lauren N. and Slagle, Adam C. and Wood, Charles D. and Zheng, Yang Yang and Gerling, Gregory J.},
  booktitle={2011 IEEE Systems and Information Engineering Design Symposium}, 
  title={Hand tracking and visualization in a virtual reality simulation}, 
  year={2011},
  volume={},
  number={},
  pages={127-132},
  doi={10.1109/SIEDS.2011.5876867}}

@INPROCEEDINGS{vr4,
author={Höll, Markus and Oberweger, Markus and Arth, Clemens and Lepetit, Vincent},
booktitle={2018 IEEE Conference on Virtual Reality and 3D User Interfaces (VR)},
title={Efficient Physics-Based Implementation for Realistic Hand-Object Interaction in Virtual Reality},
year={2018},
volume={},
number={},
pages={175-182},
doi={10.1109/VR.2018.8448284}}

@INPROCEEDINGS{sign1,
  author={Isaacs, J. and Foo, S.},
  booktitle={Thirty-Sixth Southeastern Symposium on System Theory, 2004. Proceedings of the}, 
  title={Hand pose estimation for American sign language recognition}, 
  year={2004},
  volume={},
  number={},
  pages={132-136},
  doi={10.1109/SSST.2004.1295634}}

@Article{sign2,
AUTHOR = {Shin, Jungpil and Matsuoka, Akitaka and Hasan, Md. Al Mehedi and Srizon, Azmain Yakin},
TITLE = {American Sign Language Alphabet Recognition by Extracting Feature from Hand Pose Estimation},
JOURNAL = {Sensors},
VOLUME = {21},
YEAR = {2021},
NUMBER = {17},
ARTICLE-NUMBER = {5856},
URL = {https://www.mdpi.com/1424-8220/21/17/5856},
PubMedID = {34502747},
ISSN = {1424-8220},
DOI = {10.3390/s21175856}
}

@ARTICLE{sign3,
author={Starner, T. and Weaver, J. and Pentland, A.},
journal={IEEE Transactions on Pattern Analysis and Machine Intelligence},
title={Real-time American sign language recognition using desk and wearable computer based video},
year={1998},
volume={20},
number={12},
pages={1371-1375},
doi={10.1109/34.735811}}

@INPROCEEDINGS {park1,
author = {M. Islam and S. Lee and A. Abdelkader and S. Park and E. Hoque},
booktitle = {2023 11th International Conference on Affective Computing and Intelligent Interaction Workshops and Demos (ACIIW)},
title = {PARK: Parkinson’s Analysis with Remote Kinetic-tasks},
year = {2023},
volume = {},
issn = {},
pages = {1-3},
doi = {10.1109/ACIIW59127.2023.10388168},
url = {https://doi.ieeecomputersociety.org/10.1109/ACIIW59127.2023.10388168},
publisher = {IEEE Computer Society},
address = {Los Alamitos, CA, USA},
month = {sep}
}

@article{park2,
  title={Using AI to measure Parkinson’s disease severity at home},
  author={Islam, Md Saiful and Rahman, Wasifur and Abdelkader, Abdelrahman and Lee, Sangwu and Yang, Phillip T and Purks, Jennifer Lynn and Adams, Jamie Lynn and Schneider, Ruth B and Dorsey, Earl Ray and Hoque, Ehsan},
  journal={npj Digital Medicine},
  volume={6},
  number={1},
  pages={156},
  year={2023},
  publisher={Nature Publishing Group UK London}
}

@article{tompson2014real,
  title={Real-time continuous pose recovery of human hands using convolutional networks},
  author={Tompson, Jonathan and Stein, Murphy and Lecun, Yann and Perlin, Ken},
  journal={ACM Transactions on Graphics (ToG)},
  volume={33},
  number={5},
  pages={1--10},
  year={2014},
  publisher={ACM New York, NY, USA}
}

@article{ita,
    author = {Del Bino, S. and Bernerd, F.},
    title = "{Variations in skin colour and the biological consequences of ultraviolet radiation exposure}",
    journal = {British Journal of Dermatology},
    volume = {169},
    number = {s3},
    pages = {33-40},
    year = {2013},
    month = {10},
    issn = {0007-0963},
    doi = {10.1111/bjd.12529},
    url = {https://doi.org/10.1111/bjd.12529},
    eprint = {https://academic.oup.com/bjd/article-pdf/169/s3/33/47525877/bjd0033.pdf},
}

@INPROCEEDINGS{deeppose,
  author={Toshev, Alexander and Szegedy, Christian},
  booktitle={2014 IEEE Conference on Computer Vision and Pattern Recognition}, 
  title={DeepPose: Human Pose Estimation via Deep Neural Networks}, 
  year={2014},
  volume={},
  number={},
  pages={1653-1660},
  doi={10.1109/CVPR.2014.214}}

@inproceedings{openpose-hand,
  author = {Tomas Simon and Hanbyul Joo and Iain Matthews and Yaser Sheikh},
  booktitle = {CVPR},
  title = {Hand Keypoint Detection in Single Images using Multiview Bootstrapping},
  year = {2017}
}

@article{10.1145/3678532,
author = {Wang, Qi and Luo, Can and Yang, Shuai and Xu, Shuo and Yang, Yibo and Jia, Jie and Yu, Bin},
title = {MRehab: A Mixed Reality Rehabilitation System Supporting Integrated Speech and Hand Training},
year = {2024},
issue_date = {August 2024},
publisher = {Association for Computing Machinery},
address = {New York, NY, USA},
volume = {8},
number = {3},
url = {https://doi.org/10.1145/3678532},
doi = {10.1145/3678532},
journal = {Proc. ACM Interact. Mob. Wearable Ubiquitous Technol.},
month = sep,
articleno = {133},
numpages = {23}
}

@article{10.1145/3569499,
author = {Tang, Xiao and Li, Ruihui and Fu, Chi-Wing},
title = {CAFI-AR: Contact-aware Freehand Interaction with AR Objects},
year = {2023},
issue_date = {December 2022},
publisher = {Association for Computing Machinery},
address = {New York, NY, USA},
volume = {6},
number = {4},
url = {https://doi.org/10.1145/3569499},
doi = {10.1145/3569499},
month = jan,
articleno = {183},
numpages = {23}
}

@article{10.1145/3610881,
author = {Li, Jiyang and Huang, Lin and Shah, Siddharth and Jones, Sean J. and Jin, Yincheng and Wang, Dingran and Russell, Adam and Choi, Seokmin and Gao, Yang and Yuan, Junsong and Jin, Zhanpeng},
title = {SignRing: Continuous American Sign Language Recognition Using IMU Rings and Virtual IMU Data},
year = {2023},
issue_date = {September 2023},
publisher = {Association for Computing Machinery},
address = {New York, NY, USA},
volume = {7},
number = {3},
url = {https://doi.org/10.1145/3610881},
doi = {10.1145/3610881},
journal = {Proc. ACM Interact. Mob. Wearable Ubiquitous Technol.},
month = sep,
articleno = {107},
numpages = {29}
}

@InProceedings{pmlr-v68-jaroensri17a,
  title = 	 {A Video-Based Method for Automatically Rating Ataxia},
  author = 	 {Jaroensri, Ronnachai and Zhao, Amy and Balakrishnan, Guha and Lo, Derek and Schmahmann, Jeremy D. and Durand, Fredo and Guttag, John},
  booktitle = 	 {Proceedings of the 2nd Machine Learning for Healthcare Conference},
  pages = 	 {204--216},
  year = 	 {2017},
  editor = 	 {Doshi-Velez, Finale and Fackler, Jim and Kale, David and Ranganath, Rajesh and Wallace, Byron and Wiens, Jenna},
  volume = 	 {68},
  series = 	 {Proceedings of Machine Learning Research},
  month = 	 {18--19 Aug},
  publisher =    {PMLR},
  url = 	 {https://proceedings.mlr.press/v68/jaroensri17a.html},
}

@inproceedings{videoplace,
author = {Krueger, Myron W. and Gionfriddo, Thomas and Hinrichsen, Katrin},
title = {VIDEOPLACE—an artificial reality},
year = {1985},
isbn = {0897911490},
publisher = {Association for Computing Machinery},
address = {New York, NY, USA},
url = {https://doi.org/10.1145/317456.317463},
doi = {10.1145/317456.317463},
booktitle = {Proceedings of the SIGCHI Conference on Human Factors in Computing Systems},
pages = {35–40},
numpages = {6},
location = {San Francisco, California, USA},
series = {CHI '85}
}

@inproceedings{put-that-there,
  title={“Put-that-there” Voice and gesture at the graphics interface},
  author={Bolt, Richard A},
  booktitle={Proceedings of the 7th annual conference on Computer graphics and interactive techniques},
  pages={262--270},
  year={1980}
}

@INPROCEEDINGS{qian2014hand,
  author={Qian, Chen and Sun, Xiao and Wei, Yichen and Tang, Xiaoou and Sun, Jian},
  booktitle={2014 IEEE Conference on Computer Vision and Pattern Recognition}, 
  title={Realtime and Robust Hand Tracking from Depth}, 
  year={2014},
  volume={},
  number={},
  pages={1106-1113},
  doi={10.1109/CVPR.2014.145}}

@inproceedings{simon2017hand,
  title={Hand keypoint detection in single images using multiview bootstrapping},
  author={Simon, Tomas and Joo, Hanbyul and Matthews, Iain and Sheikh, Yaser},
  booktitle={Proceedings of the IEEE conference on Computer Vision and Pattern Recognition},
  pages={1145--1153},
  year={2017}
}

@INPROCEEDINGS{sinha2016deephand,
  author={Sinha, Ayan and Choi, Chiho and Ramani, Karthik},
  booktitle={2016 IEEE Conference on Computer Vision and Pattern Recognition (CVPR)}, 
  title={DeepHand: Robust Hand Pose Estimation by Completing a Matrix Imputed with Deep Features}, 
  year={2016},
  volume={},
  number={},
  pages={4150-4158},
  doi={10.1109/CVPR.2016.450}}

@INPROCEEDINGS{mueller2018ganerated,
  author={Mueller, Franziska and Bernard, Florian and Sotnychenko, Oleksandr and Mehta, Dushyant and Sridhar, Srinath and Casas, Dan and Theobalt, Christian},
  booktitle={2018 IEEE/CVF Conference on Computer Vision and Pattern Recognition}, 
  title={GANerated Hands for Real-Time 3D Hand Tracking from Monocular RGB}, 
  year={2018},
  volume={},
  number={},
  pages={49-59},
  doi={10.1109/CVPR.2018.00013}}

@article{chua2024motion,
  title={Motion as Emotion: Detecting Affect and Cognitive Load from Free-Hand Gestures in VR},
  author={Chua, Phoebe and Sasikumar, Prasanth and Weerasinghe, Yadeesha and Nanayakkara, Suranga},
  journal={arXiv preprint arXiv:2409.12921},
  year={2024}
}

@article{BLYTHE2023105260,
title = {Emotion is perceived accurately from isolated body parts, especially hands},
journal = {Cognition},
volume = {230},
pages = {105260},
year = {2023},
issn = {0010-0277},
doi = {https://doi.org/10.1016/j.cognition.2022.105260},
url = {https://www.sciencedirect.com/science/article/pii/S0010027722002487},
author = {Ellen Blythe and Lúcia Garrido and Matthew R. Longo}
}

@article{levy,
   author = "Levy, Rachel S. and Kelly, Spencer D.",
   title = "Emotion matters", 
   journal= "Gesture",
   year = "2020",
   volume = "19",
   number = "1",
   pages = "41-71",
   doi = "https://doi.org/10.1075/gest.19029.lev",
   url = "https://www.jbe-platform.com/content/journals/10.1075/gest.19029.lev",
   publisher = "John Benjamins",
   issn = "1568-1475",
   type = "Journal Article"
  }

@incollection{KRAUSS1996389,
title = {Nonverbal Behavior and Nonverbal Communication: What do Conversational Hand Gestures Tell Us?},
editor = {Mark P. Zanna},
series = {Advances in Experimental Social Psychology},
publisher = {Academic Press},
volume = {28},
pages = {389-450},
year = {1996},
issn = {0065-2601},
doi = {https://doi.org/10.1016/S0065-2601(08)60241-5},
url = {https://www.sciencedirect.com/science/article/pii/S0065260108602415},
author = {Robert M. Krauss and Yihsiu Chen and Purnima Chawla}
}

@article{kelly,
author = {Kelly, Spencer D. and Ngo Tran, Quang-Anh},
title = {Exploring the Emotional Functions of Co-Speech Hand Gesture in Language and Communication},
journal = {Topics in Cognitive Science},
volume = {n/a},
number = {n/a},
pages = {},
doi = {https://doi.org/10.1111/tops.12657},
url = {https://onlinelibrary.wiley.com/doi/abs/10.1111/tops.12657},
eprint = {https://onlinelibrary.wiley.com/doi/pdf/10.1111/tops.12657},
}

@inproceedings{luo,
author = {Luo, Yuhan and Yu, Junnan and Liang, Minhui and Wan, Yichen and Zhu, Kening and Santosa, Shannon Sie},
title = {Emotion Embodied: Unveiling the Expressive Potential of Single-Hand Gestures},
year = {2024},
isbn = {9798400703300},
publisher = {Association for Computing Machinery},
address = {New York, NY, USA},
url = {https://doi.org/10.1145/3613904.3642255},
doi = {10.1145/3613904.3642255},
booktitle = {Proceedings of the 2024 CHI Conference on Human Factors in Computing Systems},
articleno = {404},
numpages = {17},
location = {Honolulu, HI, USA},
series = {CHI '24}
}

@article{hand-over-face,
author = {Mahmoud, Marwa and Baltru\v{s}aitis, Tadas and Robinson, Peter},
title = {Automatic Analysis of Naturalistic Hand-Over-Face Gestures},
year = {2016},
issue_date = {August 2016},
publisher = {Association for Computing Machinery},
address = {New York, NY, USA},
volume = {6},
number = {2},
issn = {2160-6455},
url = {https://doi.org/10.1145/2946796},
doi = {10.1145/2946796},
journal = {ACM Trans. Interact. Intell. Syst.},
month = jul,
articleno = {19},
numpages = {18}
}


\newpage
\newpage
\appendix

\onecolumn

\section{Appendix}


\begin{figure*}
    \centering
    \includegraphics[width=0.6\linewidth]{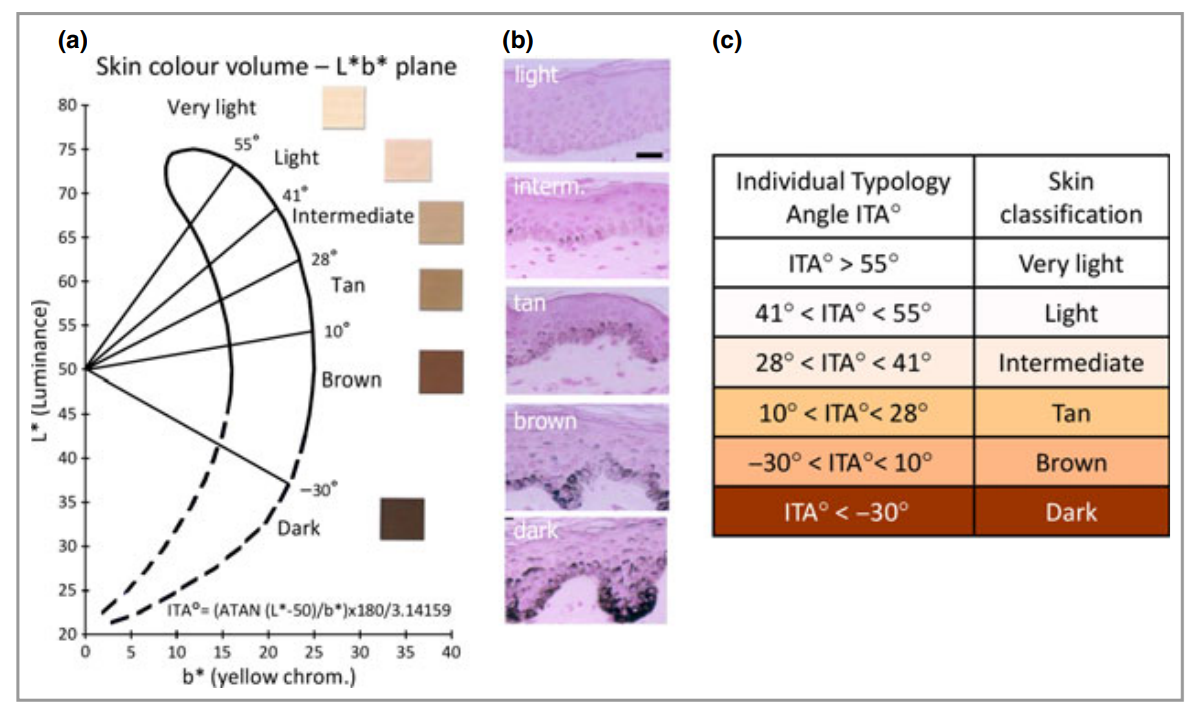}
    \caption{ITA scale used in dermatology literature to determine Ultraviolet ray exposure on human skin. Image licensed from \cite{ita}.}
    \label{fig:ita-range}
\end{figure*}

\begin{table*}[ht]
\centering
\caption{Distribution of the data augmentation techniques. Superscript \textit{I} indicates that the augmentation was selected independently. Total augmentation calculation excludes the flip operations.}
\begin{tabularx}{\textwidth}{>{\raggedright}p{0.35\textwidth}|>{\raggedright}p{0.35\textwidth}|>{\raggedright\arraybackslash}p{0.2\textwidth}}
\toprule
\textbf{Category} & \textbf{Technique} & \textbf{Percentage} \\
\midrule
\multirow{4}{=}{Geometric Transformations\textsuperscript{I} (30\%)} 
 & Downscale/Upscale & 7.50\% \\
 & Scale & 7.50\% \\
 & Stretch & 7.50\% \\
 & Translate & 7.50\% \\
\midrule
\multirow{9}{=}{Color Space Operations\textsuperscript{I} (30\%)} 
 & Brightness  & 3.33\% \\
 & Color Balance & 3.33\% \\
 & Contrast & 3.33\% \\
 & Equalize & 3.33\% \\
 & Kernel Filter & 3.33\% \\
 & Noise Injection & 3.33\% \\
 & Patch Shuffle & 3.33\% \\
 & Solarize & 3.33\% \\
 & Solarize Add & 3.33\% \\
\midrule
\multirow{4}{=}{Other Augmentations}
 & Blur\textsuperscript{I} & 50.00\% \\
 & Vertical Flip\textsuperscript{I} & 50.00\% \\
 & Horizontal Flip\textsuperscript{I} & 50.00\% \\
 & Gaussian Erase\textsuperscript{I} & 15.00\% \\
\midrule
\textbf{At least one augmentation applied} & & \textbf{79.18\%} \\
\bottomrule
\end{tabularx}
\label{tab:augmentation}
\end{table*}

\begin{figure*}[ht]
    \centering
    \begin{tabular}{*{5}{@{\hspace{2mm}}m{0.18\textwidth}@{}}}
        \includegraphics[width=\linewidth]{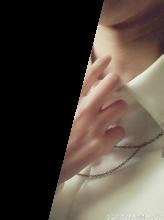} & 
        \includegraphics[width=\linewidth]{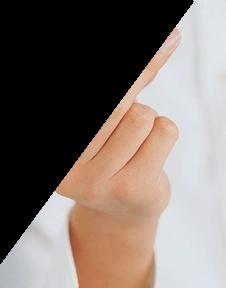} & 
        \includegraphics[width=\linewidth]{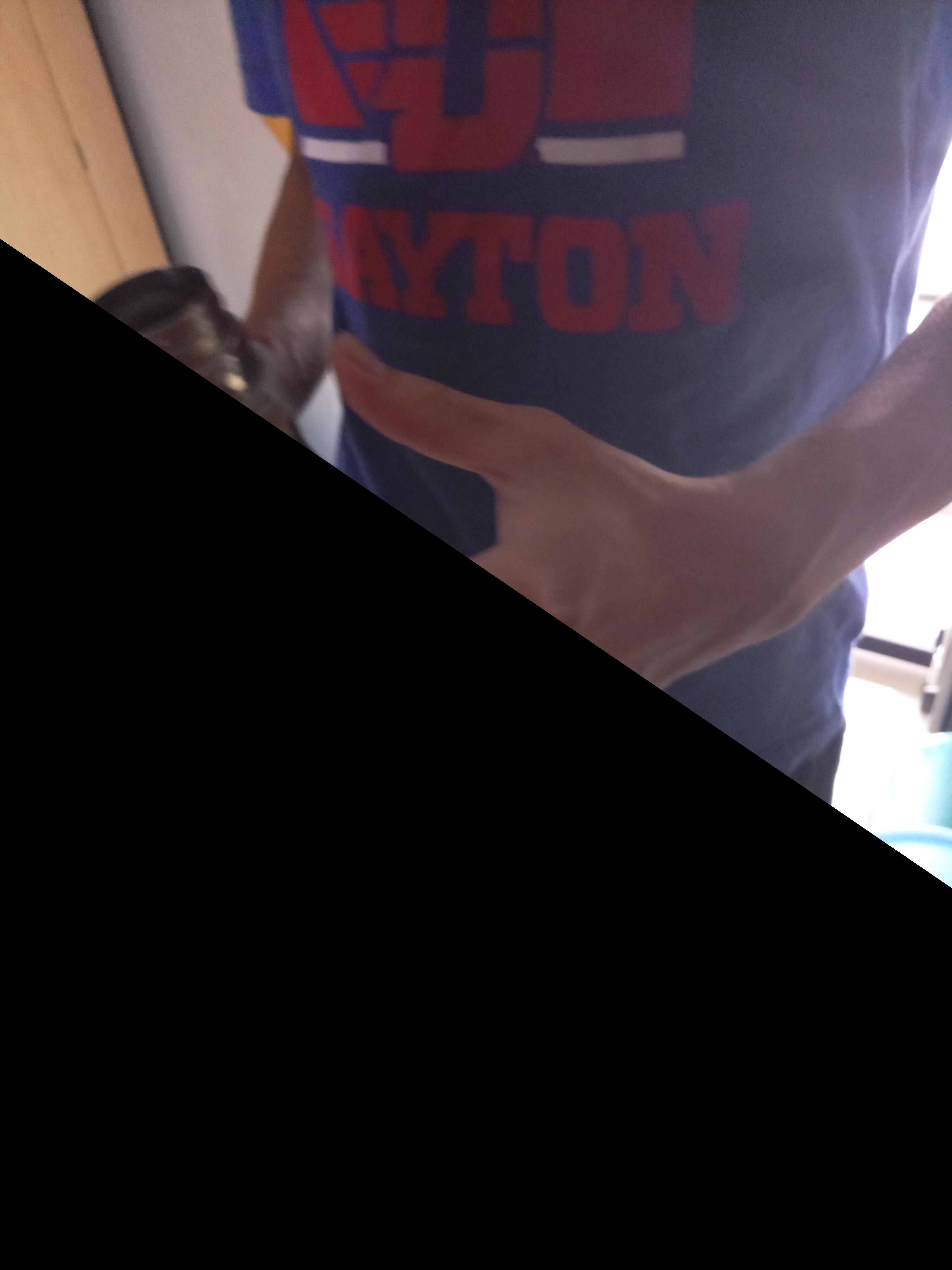} & 
        \includegraphics[width=\linewidth]{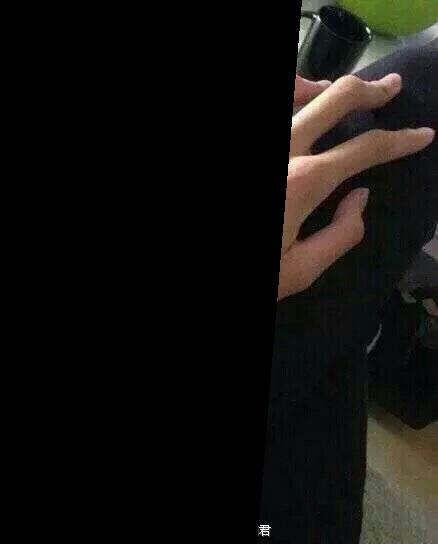} &
        \includegraphics[width=\linewidth]{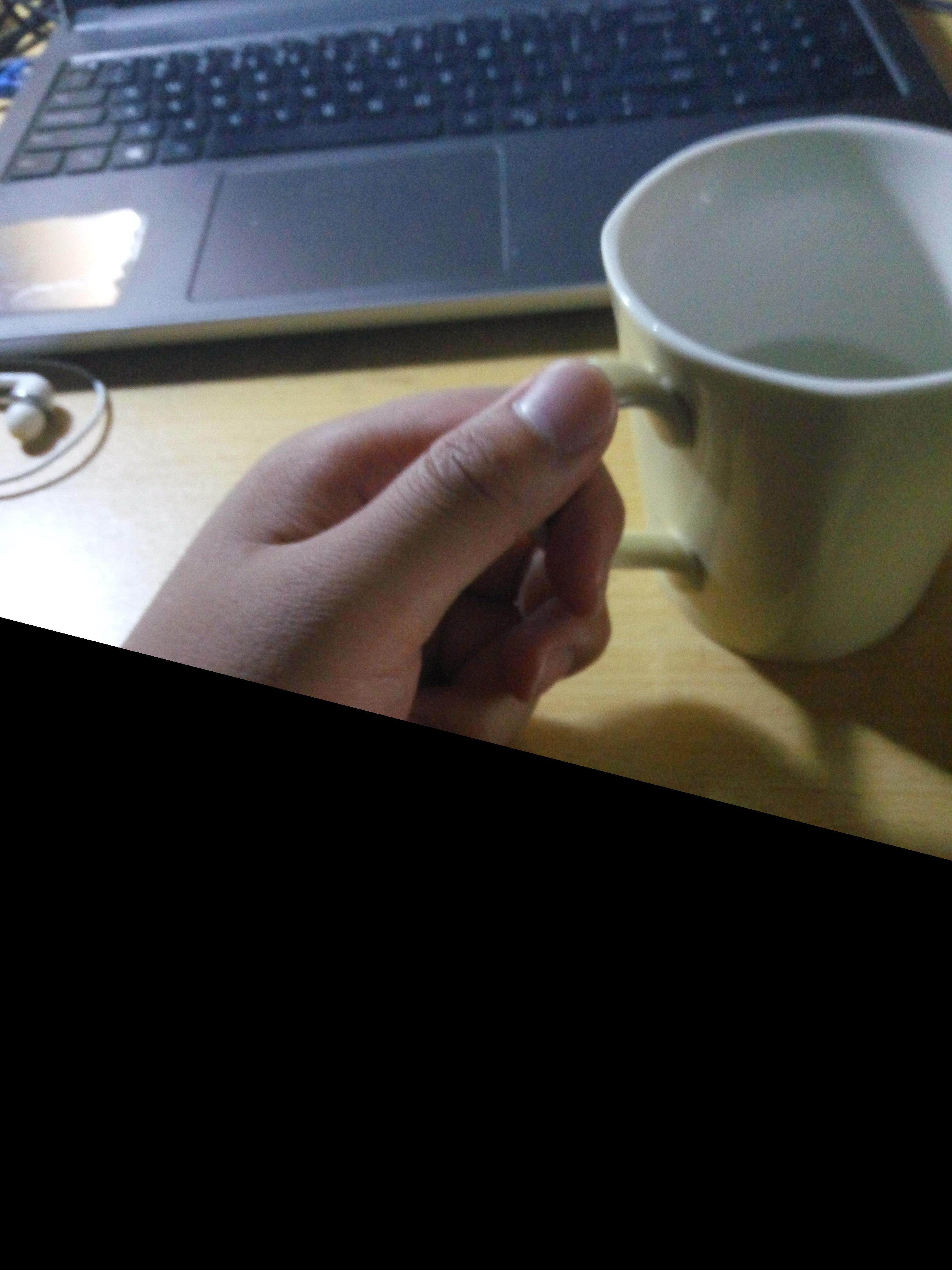} \\[5pt]
    \end{tabular}
    \caption{Sample images from perturbation test, where half of the hand in each image is hidden.}
    \label{fig:perturbed}
\end{figure*}

\begin{figure*}[ht!]
    \centering
    \begin{minipage}{0.24\textwidth}
        \centering
        \includegraphics[width=\textwidth]{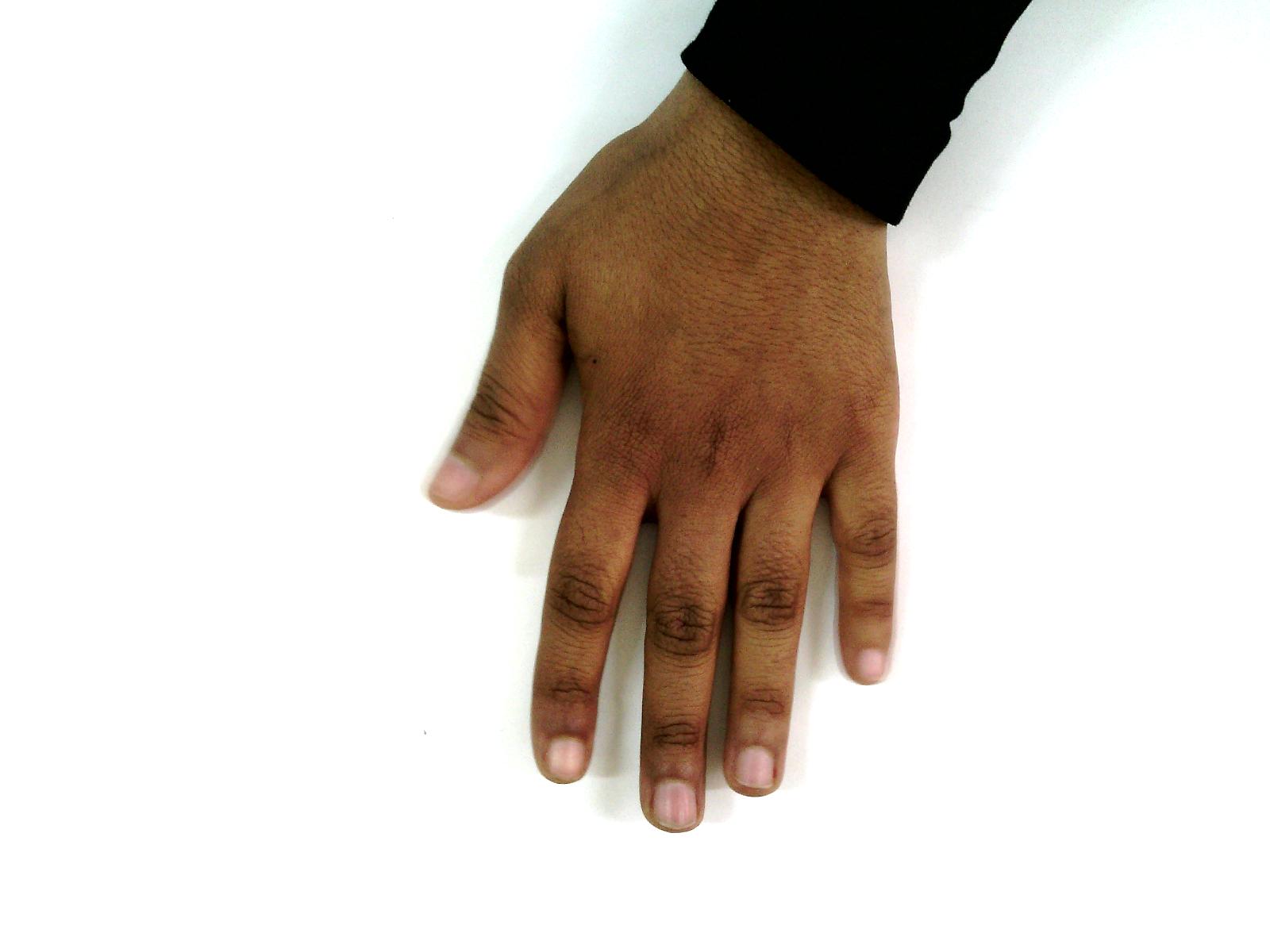}
        \caption*{Dark}
    \end{minipage}\hfill
    \begin{minipage}{0.24\textwidth}
        \centering
        \includegraphics[width=\textwidth]{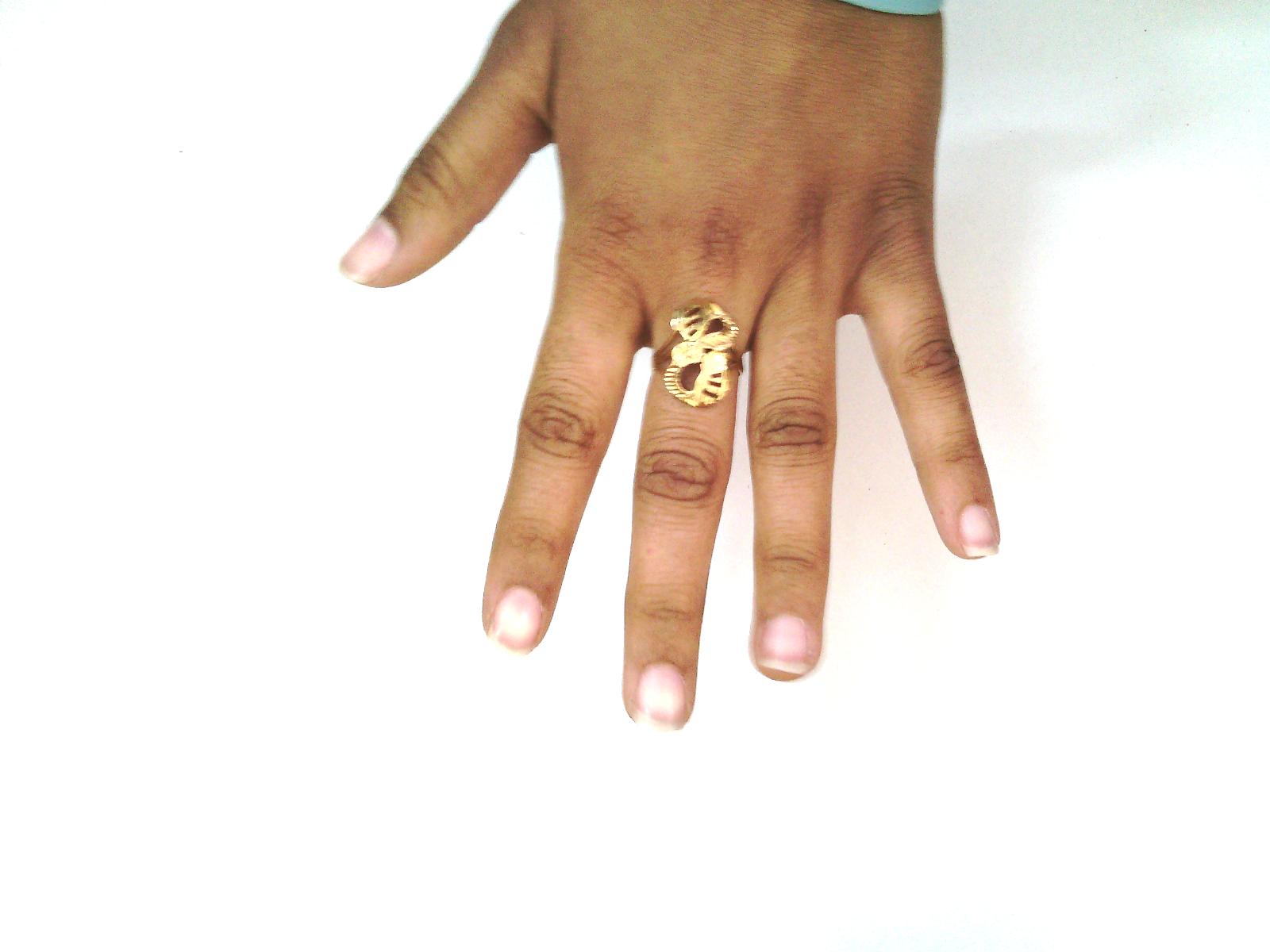}
        \caption*{Medium}
    \end{minipage}\hfill
    \begin{minipage}{0.24\textwidth}
        \centering
        \includegraphics[width=\textwidth]{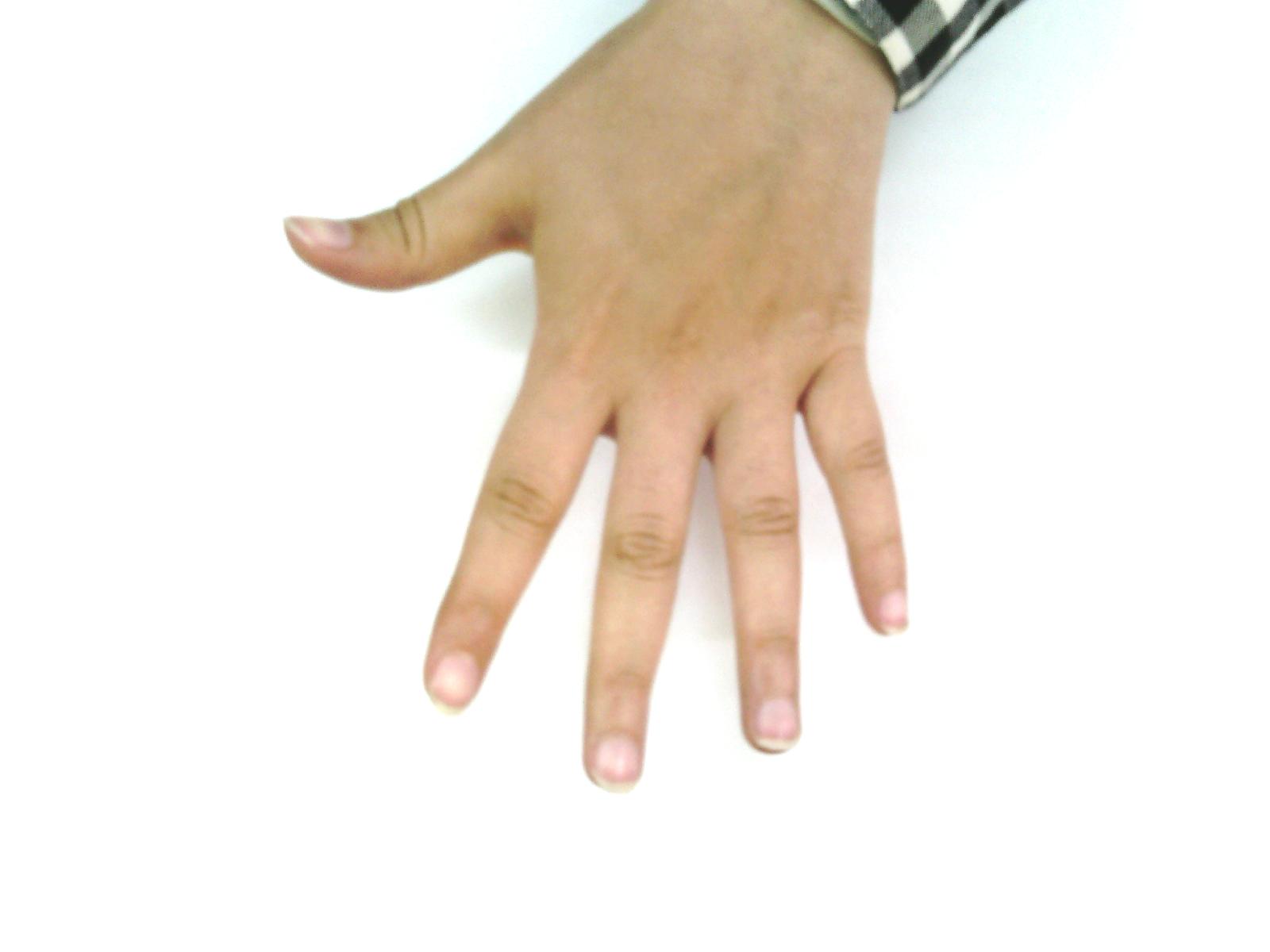}
        \caption*{Fair}
    \end{minipage}\hfill
    \begin{minipage}{0.24\textwidth}
        \centering
        \includegraphics[width=\textwidth]{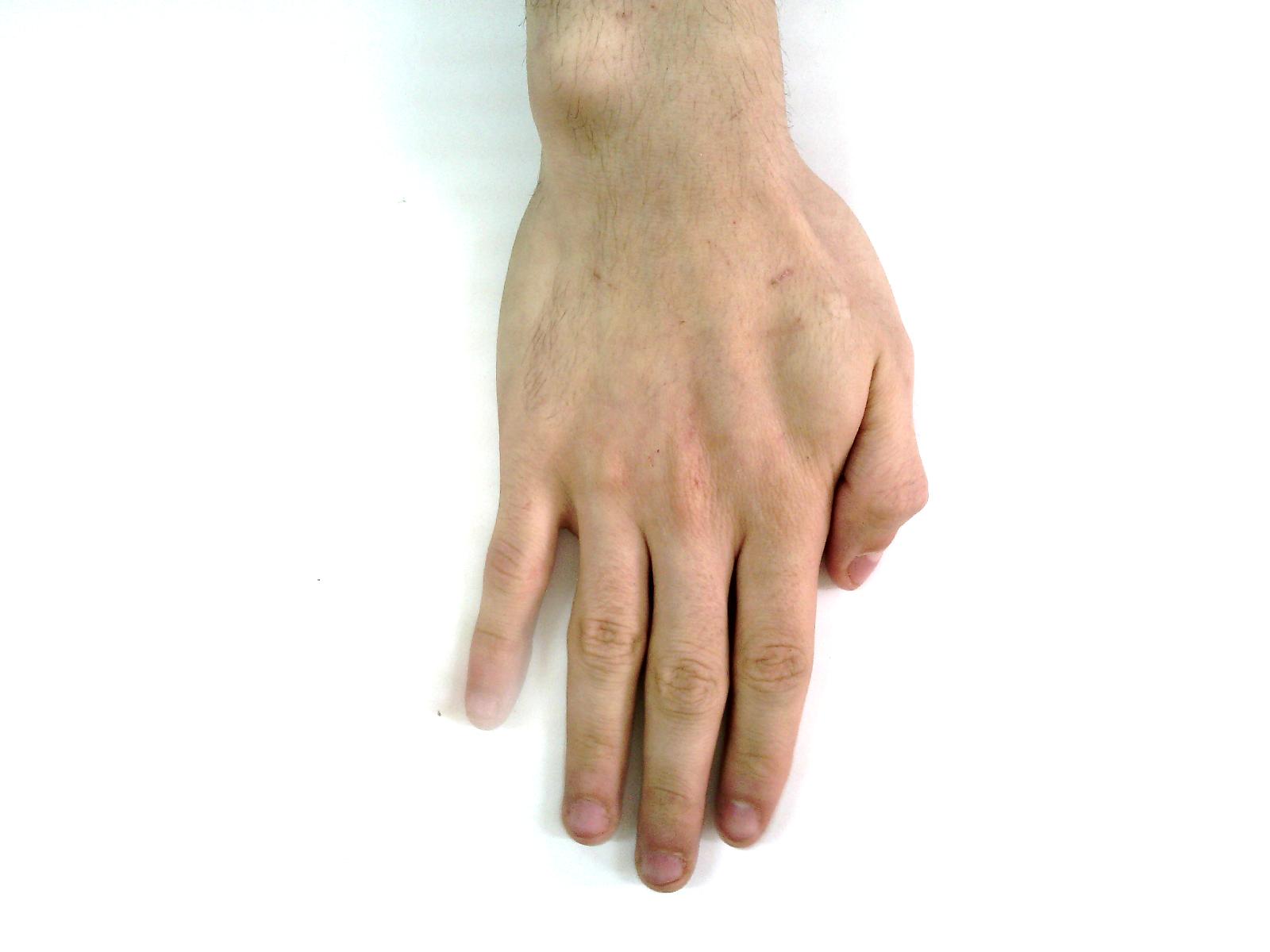}
        \caption*{Very fair}
    \end{minipage}
    \caption{4 different skin colors in 11K Hands dataset\cite{11kHands}.}
    \label{fig:11k-skin-color}
\end{figure*}

\end{document}